\documentclass{article}

\usepackage[final]{neurips_2025}

\usepackage[utf8]{inputenc} 
\usepackage[T1]{fontenc}    
\usepackage{hyperref}       
\usepackage{url}            
\usepackage{booktabs}       
\usepackage{amsfonts}       
\usepackage{nicefrac}       
\usepackage{microtype}      
\usepackage{xcolor}         

\usepackage{graphicx} 
\usepackage{amsmath, amssymb, algorithm, algpseudocode}
\usepackage{algorithmicx}
\usepackage{algpseudocode}
\usepackage{algorithm}
\usepackage{booktabs}
\usepackage{multirow} 
\usepackage{enumitem}

\usepackage{caption}  

\usepackage{wrapfig}

\title{Communication-Efficient Diffusion Denoising Parallelization via Reuse-then-Predict Mechanism}

\author{%
  Kunyun Wang\thanks{Equal contribution.} \\
  School of Computer Science,\\ Shanghai Jiao Tong University \\
  \texttt{wkykaixin@sjtu.edu.cn} \\
  \And
  Bohan Li\footnotemark[1] \\
  School of Computer Science,\\ Shanghai Jiao Tong University \\
  \texttt{everlastingnight@sjtu.edu.cn} \\
  \And
  Kai Yu \\
  School of Computer Science,\\ Shanghai Jiao Tong University \\
  \texttt{kai.yu@sjtu.edu.cn} \\
  \And
  Minyi Guo \\
  School of Computer Science,\\ Shanghai Jiao Tong University \\
  \texttt{guo-my@cs.sjtu.edu.cn} \\
  \And
  Jieru Zhao\thanks{Corresponding author.} \\
  School of Computer Science,\\ Shanghai Jiao Tong University \\
  \texttt{zhao-jieru@sjtu.edu.cn} \\
}

\begin{document}

\maketitle

\begin{abstract}
Diffusion models have emerged as a powerful class of generative models across various modalities, including image, video, and audio synthesis. However, their deployment is often limited by significant inference latency, primarily due to the inherently sequential nature of the denoising process. While existing parallelization strategies attempt to accelerate inference by distributing computation across multiple devices, they typically incur high communication overhead, hindering deployment on commercial hardware. To address this challenge, we propose \textbf{ParaStep}, a novel parallelization method based on a reuse-then-predict mechanism that parallelizes diffusion inference by exploiting similarity between adjacent denoising steps. Unlike prior approaches that rely on layer-wise or stage-wise communication, ParaStep employs lightweight, step-wise communication, substantially reducing overhead. ParaStep achieves end-to-end speedups of up to \textbf{3.88}$\times$ on SVD, \textbf{2.43}$\times$ on CogVideoX-2b, and \textbf{6.56}$\times$ on AudioLDM2-large, while maintaining generation quality. These results highlight ParaStep as a scalable and communication-efficient solution for accelerating diffusion inference, particularly in bandwidth-constrained environments.

\end{abstract}

\newcommand{\LBH}[2]{\textcolor{red}{(Bohan: #1})}
\section{Introduction}
\label{sec:Intro}
Benefiting from advances in deep learning and increasingly powerful hardware, diffusion models have demonstrated remarkable performance in image generation \cite{image1,image2,image3,flux2024,hunyuandit}, video generation \cite{Cogvideox, Hunyuanvideo, Latte, SVD, consisid}, and audio generation \cite{audio1,audio2,audio3,audioldm, Audioldm2}. However, their widespread adoption is severely limited by substantial inference latency. This issue becomes particularly critical in long-form video generation, where producing a few minutes of video may require hours of GPU computation.

The latency primarily stems from two factors: the high computational cost of the noise predictor—typically implemented using architectures such as DiT \cite{dit} or U-Net \cite{unet}—and the inherently sequential nature of the denoising process, which invokes the noise predictor repeatedly across dozens or even hundreds of timesteps. Notably, in DiT models with 3D attention, this cost scales quadratically with both spatial resolution and temporal length.

To reduce the inference latency of diffusion models, researchers have proposed a variety of solutions. 
Distillation techniques \cite{meng2023distillation, sauer2024distillation} can reduce the computational and memory overhead of diffusion models. 
Other works \cite{qdiffusion, shang2023post} employ post-training quantization (PTQ) to compress full-precision models into 8-bit or 4-bit representations without retraining, thereby reducing computation.
Caching-based methods have also been explored to eliminate redundant computation during the denoising process \cite{deepcache, approximate, morecite1, morecite2}. While effective, these approaches do not leverage the potential of distributed computing, which has become a cornerstone of modern deep learning systems.

To address this, several works have proposed parallelization strategies that leverage distributed infrastructure to accelerate diffusion models \cite{distrifusion, Asyncdiff, pipefusion, xDiT}. These methods perform operation-wise, layer-wise, or stage-wise communication to exchange essential tensors, which incurs substantial communication cost. Despite their effectiveness, the huge communication overhead makes them impractical outside of high-bandwidth, data center–scale environments. Therefore, there remains a pressing need for parallel approaches with minimal communication requirements to enable practical deployment on commercial hardware.

\begin{figure} 
  \centering
  \includegraphics[width=\linewidth]{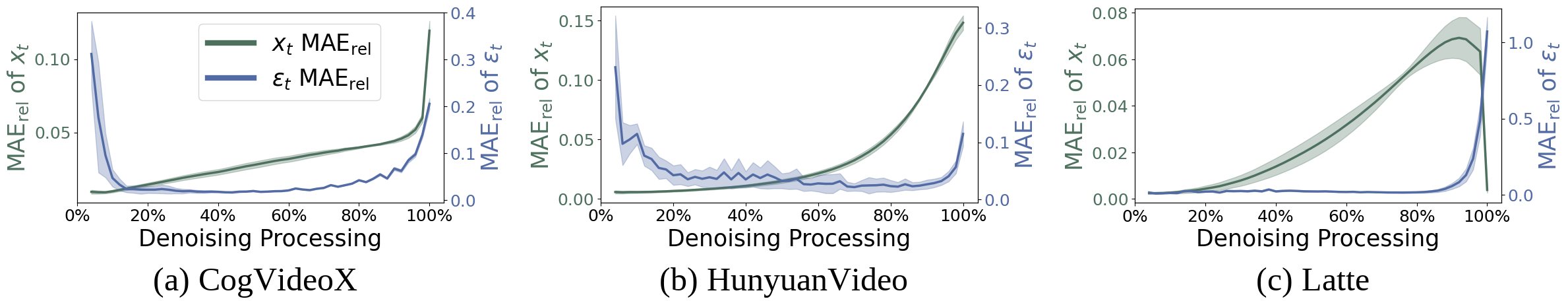}
  \vspace{-13pt}
  \caption{
  Relative MAE (as defined in Equation~\eqref{eq:rel_mae}) between adjacent denoising steps $t$ and $t+1$ for the noisy sample $\mathbf{x}_t$ and the predicted noise $\boldsymbol{\epsilon}_t$. Here, 0\% on the x-axis indicates the first step of the denoising process, and 100\% indicates the last.
  }
  \label{fig:similarity_between_steps}
\end{figure}

\begin{figure} 
  \vspace{-10pt}
  \centering
  \includegraphics[width=\linewidth]{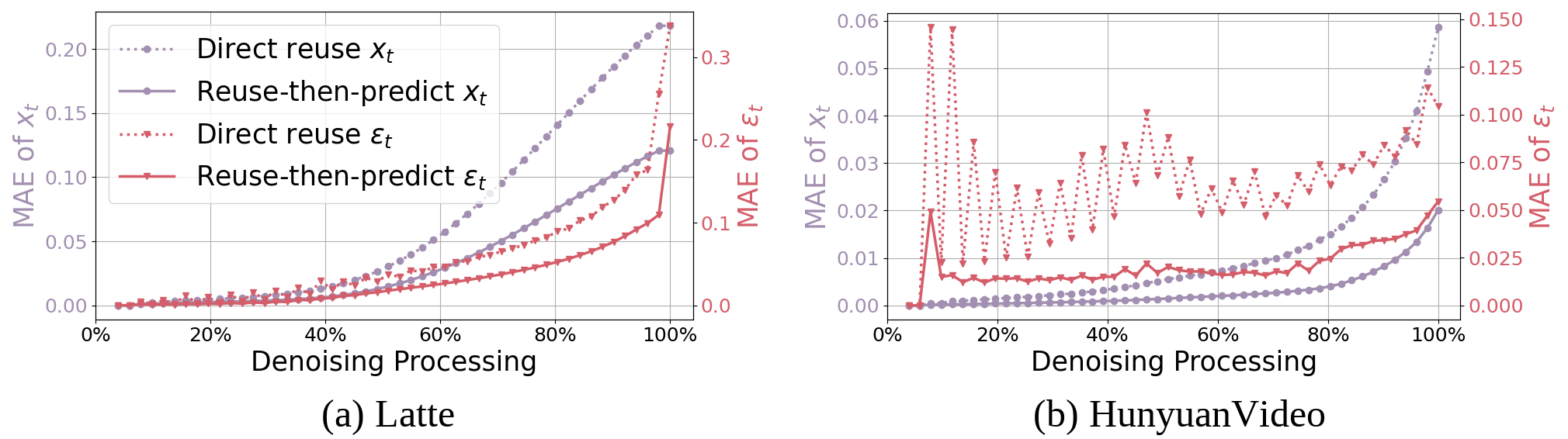}
  \vspace{-12pt}
  \caption{
  Dotted lines show the difference in noisy sample $\mathbf{x}_t$ between the original model and direct reuse process (stride = 2), while solid lines compare the original model with our reuse-then-predict process (degree = 2). Results for predicted noise $\boldsymbol{\epsilon}_t$ follow the same pattern. Reuse-then-predict process results in smaller deviations from the original model compared to direct reuse process. }
  \label{fig:diff_reuse_and_step}
\end{figure}

We analyze the differences in noisy samples and predicted noise across adjacent denoising steps. As illustrated in Figure~\ref{fig:similarity_between_steps}, except for the initial and final few steps, the predicted noises across neighboring steps exhibit high similarity. This observation suggests that it is possible to reuse the noise generated in the previous step to skip the computation of the current step, as shown in Figure~\ref{fig:reuse_and_ParaStep}(b). However, such direct reuse leads to noticeable degradation in generation quality.

To address this issue, we propose a \textit{reuse-then-predict} mechanism. Specifically, the noise from the previous step is reused to generate the noisy sample for the current step, and then a fresh noise prediction is performed based on this sample. Owing to the robustness of noise predictors, the newly predicted noise better approximates the original noise than direct reuse, as evidenced in Figure~\ref{fig:diff_reuse_and_step}. This procedure can be parallelized through careful scheduling, illustrated in Figure~\ref{fig:reuse_and_ParaStep}(c), enabling acceleration without significant loss in generation quality. Communication is required only once per denoising step, and only the samples and noises need to be transmitted. This low communication overhead makes ParaStep highly suitable for low-bandwidth environments, such as systems connected via PCIe Gen3. A detailed analysis of the communication pattern is provided in Section~\ref{subsec:communitation}.

Since the difference between adjacent timesteps is large in the early stages, and errors introduced by reuse can accumulate over multiple steps, it is necessary to perform the original denoising process during this phase before switching to the reuse-then-predict mechanism. We refer to these initial steps as \textit{warm-up} steps. The number of warm-up steps is a critical hyperparameter: setting it too low may cause significant performance degradation, while setting it too high can limit the achievable speedup. We will analyze the impact of warm-up step selection in Section~\ref{subsec:Ablation}. Although the difference between adjacent timesteps is also large in the final few steps, the errors introduced by reuse only accumulate over a small number of steps and do not influence generation performance. Therefore, we continue to apply the reuse-then-predict mechanism during this phase to maintain acceleration.

For non-compute-intensive models, such as audio diffusion models, distributed computing may be unnecessary. In such cases, enabling ParaStep on a single device can achieve acceleration with lower computational cost. Based on this insight, we develop a single-device variant of ParaStep, called BatchStep, which performs adjacent noise predictions within a single batch rather than distributing them across multiple devices. This design reduces computational overhead while preserving the acceleration benefits of our approach for lightweight models. However, for compute-intensive models such as image and video diffusion, BatchStep is not suitable due to their high resource demands. Our code is available at \href{https://github.com/sjtu-zhao-lab/ParaStep}{https://github.com/sjtu-zhao-lab/ParaStep}.

We summarize our key contributions as follows:
\begin{itemize}[leftmargin=2em]

  \item 
  We propose a reuse-then-predict mechanism that mitigates the quality degradation caused by direct reuse. Building on this mechanism, we introduce a novel distributed sampling method, termed \textbf{ParaStep}, which enables step-wise parallelization across devices for faster inference.

  \item We analyze the communication characteristics of ParaStep and demonstrate that it achieves significant speedup even under low-bandwidth settings.

  \item We extend ParaStep to a single-device variant tailored for non-compute-intensive diffusion models, enabling efficient adjacent-step prediction without cross-device communication.

  \item We perform comprehensive experiments in various diffusion-based models, demonstrating that \textbf{ParaStep} is an effective and generalized method applicable to both vision and audio modalities.
\end{itemize}

\section{Related works}
\paragraph{Cache}
Several works utilize caching mechanisms to reduce the computational cost of the noise predictor \cite{deepcache, approximate}. DeepCache \cite{deepcache} leverages the structural properties of the U-Net architecture, specifically its skip connections between shallow and deep blocks. During less critical timesteps, DeepCache skips the computation of deep blocks by directly forwarding the output of shallow blocks to their corresponding counterparts in deeper layers. In a different direction, \cite{approximate} explores caching based on prompt similarity. It constructs a cache that stores intermediate features of previously processed prompts, enabling reuse when similar prompts are encountered. While both approaches reduce the computation of the noise predictor, neither leverages the benefits of distributed computation.


\paragraph{Parallelism}
Conventional parallel strategies such as data parallelism, pipeline parallelism, and tensor parallelism are generally unsuitable for reducing the latency of diffusion models. Data parallelism and pipeline parallelism are primarily designed to improve throughput and provide little benefit for latency reduction. While tensor parallelism is effective for accelerating large language models (LLMs), it is less suitable for diffusion models due to their large activation sizes.

To fully exploit the computational power of distributed GPUs, several methods have been specifically designed to leverage the intrinsic properties of diffusion models for acceleration \cite{distrifusion, Asyncdiff, pipefusion, xDiT}. Given a parallelism degree of $p$, DistriFusion \cite{distrifusion} partitions the input latent of the noise predictor into $p$ patches and employs all-to-all communication to merge them before each attention and convolution operation. PipeFusion \cite{pipefusion} reduces communication costs by reusing one-step stale feature maps in patch-level parallelism. AsyncDiff \cite{Asyncdiff} divides the noise predictor into $p$ stages, assigning each stage to a different device and employing pipeline parallelism to achieve speedup. xDiT \cite{xDiT} further combines PipeFusion with Ring Attention \cite{ringattention} to achieve significant acceleration. However, the high communication overhead associated with these methods restricts their effectiveness to high-bandwidth environments, which are typically found only in expensive data center infrastructures. Moreover, except for Ring Attention \cite{ringattention}, all of these approaches rely on approximate parallelism methods, which introduce noticeable deviations in the generated results compared to the original model.

\paragraph{Reuse}
TeaCache \cite{TeaCache} proposes a selective mechanism to reuse the generated noise from the previous timestep, thereby skipping the computation required for the relatively unimportant timesteps. It achieves acceleration with only minor degradation in generation quality. The key difference between TeaCache and our approach lies in their focus: TeaCache addresses the question of \emph{when} to reuse, while our method focuses on \emph{how} to reuse. Specifically, our reuse-then-predict mechanism yields a smaller performance drop by predicting a refined noise estimate rather than directly reusing previous outputs. Moreover, we demonstrate that TeaCache can be seamlessly combined with our method to further improve generation quality while maintaining the same level of acceleration.

\section{Preliminaries}

Diffusion models are a class of generative models that progressively transform data from the original distribution into a Gaussian distribution through a forward diffusion process, and then reconstruct the original data from Gaussian noise via a reverse (backward) denoising process. 

\paragraph{Forward diffusion process}  
Given a data sample $\mathbf{x}_0 \sim q(x)$ drawn from the original distribution, Gaussian noise is gradually added over $T$ steps to produce the final sample $\mathbf{x}_T$, which is a pure Gaussian noise. To avoid simulating the full forward trajectory step by step, $\mathbf{x}_t$ can be sampled via:
\begin{equation}
q(\mathbf{x}_t \mid \mathbf{x}_0) = \mathcal{N} \left( \mathbf{x}_t; \sqrt{\bar{\alpha}_t} \mathbf{x}_0, (1 - \bar{\alpha}_t) \mathbf{I} \right)
\end{equation}

\paragraph{Backward diffusion process}  
Given $\mathbf{x}_t$, the sample $\mathbf{x}_{t-1}$ is drawn from the conditional distribution:
\begin{equation}
p_{\theta}(\mathbf{x}_{t-1} \mid \mathbf{x}_t) = \mathcal{N} \left( \mathbf{x}_{t-1}; \mu_{\theta}(\mathbf{x}_t, t), \sigma_t^2 \mathbf{I} \right)
\end{equation}
where $\sigma_t$ is either predefined or learned and the mean $\mu_{\theta}(\mathbf{x}_t, t)$ is computed as:
\begin{equation}
\mu_{\theta}(\mathbf{x}_t, t) = \frac{1}{\sqrt{\alpha_t}} \left( \mathbf{x}_t - \frac{1 - \alpha_t}{\sqrt{1 - \bar{\alpha}_t}} \epsilon_t \right)
\end{equation}

\paragraph{Denoising computation}  
Given a condition $c$, a timestep $t$, and a noisy sample $\mathbf{x}_t$, the noise predictor $\epsilon_{\theta}$—typically implemented as a U-Net \cite{unet} or DiT \cite{dit}—estimates the noise $\epsilon_t$. A scheduler then uses this noise to sample $\mathbf{x}_{t-1}$:
\begin{equation}
\epsilon_t = \epsilon_{\theta}(\mathbf{x}_t, t, c), \quad \mathbf{x}_{t-1} = \textbf{Scheduler}(\mathbf{x}_t, t, \epsilon_t)
\end{equation}

\paragraph{Training}  
In DDPM \cite{ddpm}, model parameters are optimized to minimize the squared error between the true noise and the predicted noise:
\begin{equation}
\nabla_{\theta} \left\| \epsilon - \epsilon_{\theta} \left( \sqrt{\bar{\alpha}_t} \mathbf{x}_0 + \sqrt{1 - \bar{\alpha}_t} \epsilon, t \right) \right\|^2
\end{equation}
where $\bar{\alpha}_t$ is derived from a predefined noise schedule $\{\beta_t\}$. At each denoising step $t$, the model aims to recover the noise originally added to $\mathbf{x}_0$ to obtain $\mathbf{x}_t$. Since the differences between adjacent noisy samples are often small, the predicted noise $\epsilon_t$ tends to vary only slightly across timesteps.

\paragraph{Flow Matching}
Flow Matching\cite{flow} is a generalization of diffusion\cite{gao2025diffusion}, sampling using the ordinary differential equation.  In Flow Matching, the generative process transforms sample $\mathbf{z}_0$ from a simple reference distribution into sample $\mathbf{z}_1$ from the target distribution by integrating a learned velocity field $\mathbf{v}_\theta(\mathbf{z}, t)$ over time. This process is discretized as:
\begin{equation}
 \mathbf{z}_{t + \Delta t} = \mathbf{z}_t + \Delta t \cdot \mathbf{v}_\theta(\mathbf{z}_t, t)
\end{equation}
 

\section{Methods}
\label{sec:method}


In conventional diffusion models, the reverse process is implemented as a sequential denoising procedure. The noise predictor $\epsilon_{\theta}$ is invoked repeatedly to predict the noise $\epsilon_t$, given inputs $\mathbf{x}_t$, $t$, and a conditioning signal $c$. For simplicity, we omit $c$ in the remainder of this paper. The scheduler then uses the predicted noise $\epsilon_t$ to compute the denoised sample $\mathbf{x}_{t-1}$. Since $\epsilon_{\theta}$ is usually a deep neural network with high computational complexity, whereas the scheduler only performs element-wise additions and multiplications, \textit{the dominant source of computational cost and latency in diffusion models lies in the repeated invocation of the noise predictor}.

To transform the Gaussian noise $\mathbf{x}_T$ into a clean sample $\mathbf{x}_0$, the noise predictor must be called sequentially for tens or even hundreds of timesteps, resulting in significant latency during inference. A naive approach to reduce this cost is to reuse the noise predicted in the previous timestep. However, such direct reuse typically leads to a noticeable degradation in generation quality. To mitigate this degradation while preserving efficiency, we propose a \textit{reuse-then-predict} mechanism, termed \textbf{ParaStep}. Our method first reuses the noise from the previous timestep to compute the current noisy sample $\mathbf{x}_t$, and then predicts a refined noise estimate $\epsilon_t$ based on that sample. Through careful design, this procedure allows the prediction of adjacent-step noises to be distributed across multiple devices, effectively parallelizing the most expensive part of the computation. Given a parallelism degree of $p$, the overall workload is divided such that each device only performs $\frac{1}{p}$ of the noise prediction steps. Before introducing the detailed implementation of ParaStep, we first analyze the similarity of predicted noise across adjacent timesteps, and then describe how previously generated noise can be efficiently reused in the denoising process.

\begin{figure} 
  \vspace{-10pt}
  \centering
  \includegraphics[width=\linewidth]{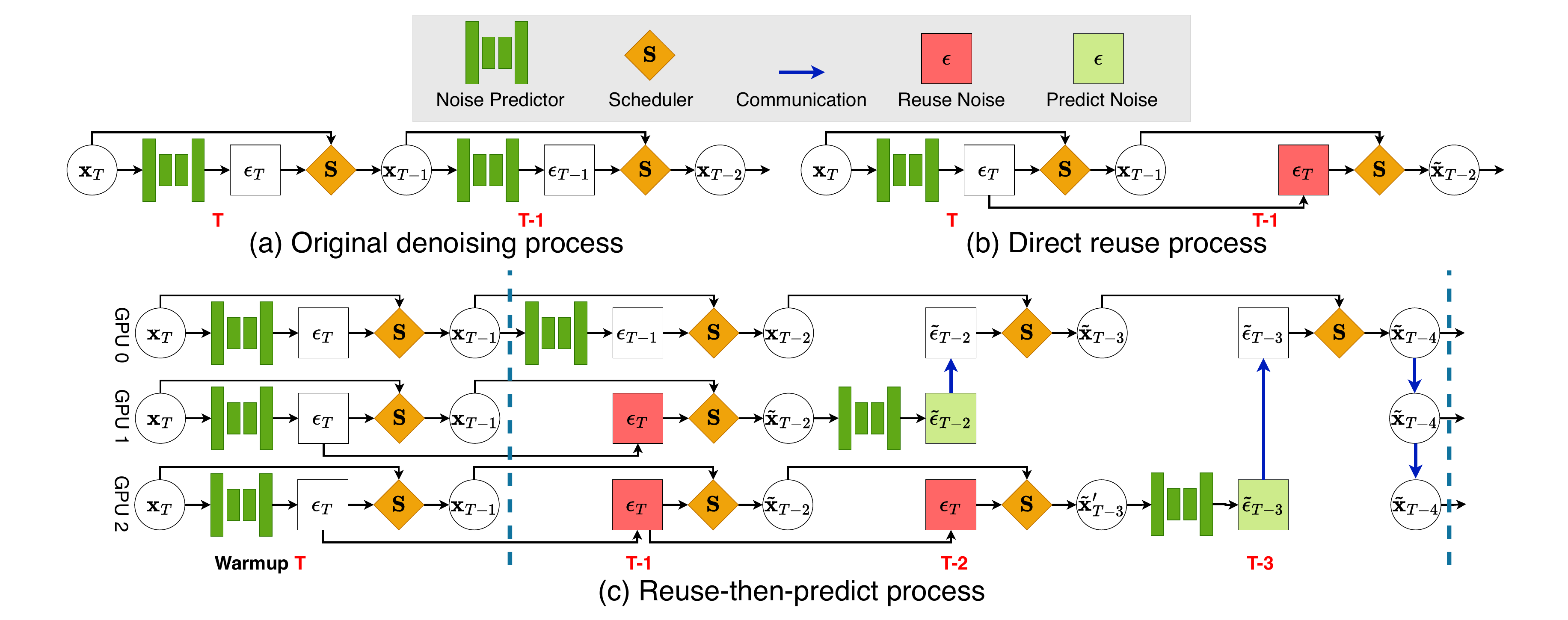}
  \vspace{-10pt}
  \caption{Illustration of the computation process of a diffusion model. (a) The original computation process. (b) Reusing noise $\epsilon_{T}$ from the previous timestep to skip the computation of noise prediction in the current timestep $T-1$. (c) ParaStep: adjacent-step noise prediction is distributed across GPUs using reuse-then-predict, enabling parallel denoising with minimal communication. Since the computational cost of scheduler operations is negligible, the noise predictor computations on GPU0, GPU1, and GPU2 are fully parallelized.
  }
  \label{fig:reuse_and_ParaStep}
\end{figure}
\subsection{Reuse-then-predict mechanism}
\paragraph{Reusing noise}
In each denoising step, the noise predictor aims to estimate the total noise that was added to $\mathbf{x}_0$ to obtain $\mathbf{x}_t$. The predicted noise $\epsilon_t$ remains highly similar across adjacent timesteps, as shown in Figure~\ref{fig:similarity_between_steps}. We use the \textit{Relative Mean Absolute Error (Relative MAE)} to quantify the similarity between noisy samples and predicted noise across steps. The metric is defined as:
\begin{equation}
\label{eq:rel_mae}
\mathrm{MAE}_{\mathrm{rel}}(\mathbf{x}_{t}, \mathbf{x}_{t+1}) =
\frac{
\frac{1}{n} \sum_{i=1}^{n} \left| x_{t,i} - x_{t+1,i} \right|
}{
\frac{1}{n} \sum_{i=1}^{n} \left| x_{t,i} \right|
}
\end{equation}

As illustrated in Figure~\ref{fig:similarity_between_steps}, for models such as CogVideoX \cite{Cogvideox} and HunyuanVideo \cite{Hunyuanvideo}, the relative MAE between adjacent noise predictions drops below 0.1 after the first 20\% of the total steps.  This observation suggests that previously predicted noise can be reused to bypass less critical computations.

In Figure~\ref{fig:reuse_and_ParaStep}(a), at timestep $T$, the noise predictor generates $\epsilon_T$ using $\mathbf{x}_T$, after which the scheduler computes $\mathbf{x}_{T-1}$ based on $\epsilon_T$, $\mathbf{x}_T$, and $T$. At timestep $T-1$, the noise predictor estimates $\epsilon_{T-1}$, which is then used to obtain $\mathbf{x}_{T-2}$. In contrast, Figure~\ref{fig:reuse_and_ParaStep}(b) illustrates a reuse strategy: the predictor first estimates $\epsilon_T$ from $\mathbf{x}_T$, which is used to obtain $\mathbf{x}_{T-1}$; then, instead of invoking the predictor again, the model reuses $\epsilon_T$ at timestep $T-1$ to estimate the next noisy sample $\tilde{\mathbf{x}}_{T-2}$. We refer to this strategy as the \textit{direct reuse process}.

Suppose the predicted noise is reused for $s-1$ subsequent timesteps. In that case, the number of predictor invocations is reduced by a factor of $s$, resulting in a corresponding $\frac{1}{s}$ reduction in latency. However, this aggressive reuse introduces a distributional mismatch from the original denoising trajectory, which can lead to significant performance degradation. To mitigate this, we propose a \textit{reuse-then-predict} mechanism that combines reuse with predictive refinement, effectively reducing the quality drop while retaining computational efficiency.

\paragraph{Parallel denoising via reuse-then-predict mechanism}

As shown in Figure~\ref{fig:diff_reuse_and_step}, instead of directly reusing the noise predicted in the previous step, a more accurate result can be achieved by first using the reused noise to generate the noisy sample for the current step, and then feeding this sample into the noise predictor to predict the refined noise for the next timestep, we call this mechanism \textit{reuse-then-predict}. Based on this mechanism, we propose a novel distributed sampling method, \textbf{ParaStep}, which parallelizes inference by distributing adjacent-step noise prediction across devices.

As illustrated in Figure~\ref{fig:reuse_and_ParaStep}(c), both the noise predictor and the scheduler are replicated on GPU0, GPU1, and GPU2. During the initial warm-up steps, all GPUs follow the standard sequential denoising process. After warm-up, GPU0 predicts $\epsilon_{T-1}$ from $\mathbf{x}_{T-1}$ and timestep $T-1$, and the scheduler then computes $\mathbf{x}_{T-2}$ accordingly. In parallel, GPU1 skips the computation of $\epsilon_{T-1}$ and instead reuses $\epsilon_T$ from the previous step to compute an approximate noisy sample $\tilde{\mathbf{x}}_{T-2}$. It then uses this sample and timestep $T-2$ to predict a refined noise estimate $\tilde{\epsilon}_{T-2}$. Simultaneously, GPU2 performs the same reuse-then-predict process, reusing $\epsilon_{T}$ twice to compute $\tilde{\mathbf{x}}^{\prime}_{T-3}$ and then predicting $\tilde{\epsilon}_{T-3}$ from $\tilde{\mathbf{x}}^{\prime}_{T-3}$ and $T-3$. Since the scheduler introduces negligible computational overhead, the noise prediction steps on GPU1 and GPU2 are effectively parallelized with those on GPU0. Once prediction is complete, GPU1 and GPU2 transmit their predicted noise values, $\tilde{\epsilon}_{T-2}$ and $\tilde{\epsilon}_{T-3}$ respectively, to GPU0. GPU0 then uses these values to compute the next set of noisy samples, $\tilde{\mathbf{x}}_{T-3}$ and $\tilde{\mathbf{x}}_{T-4}$. Finally, the newly generated sample $\tilde{\mathbf{x}}_{T-4}$ is broadcast back to GPU1 and GPU2, and all GPUs advance to timestep $T-4$. This cycle is repeated for subsequent steps, enabling efficient parallelization of adjacent-step noise prediction across multiple devices.  We formalize ParaStep with a round-based algorithm, presented in Appendix \ref{sec:Implementation}.

Assuming a parallelism degree of $p$, ParaStep operates in cycles of length $p$. Each GPU performs one forward pass of the noise predictor per cycle. Since scheduler operations are negligible in cost, noise prediction across all devices is effectively parallelized. Consequently, the total latency of the denoising process is reduced by a factor of $\frac{1}{p}$, achieving a theoretical speedup of ${p}$.




\subsection{Communication analysis}
\label{subsec:communitation}
\begin{figure} 
  \vspace{-12pt}
  \centering
  \includegraphics[width=\linewidth]{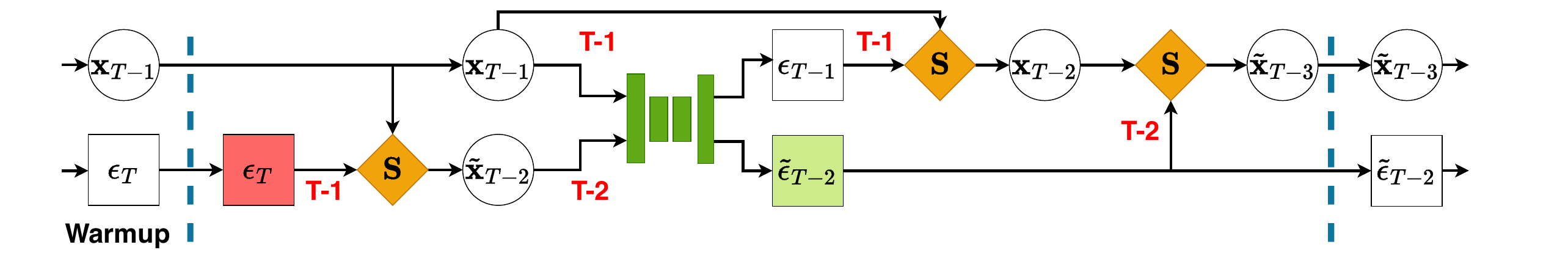}
  \vspace{-10pt}
  \caption{Utilizing batching effect of audio diffusion models, parallel computation of noise predictor can be transformed batching execution on single device. }
  \label{fig:local_batching}
\end{figure}
To evaluate the communication overhead in multi-GPU environments, we compare the \textbf{total} communication volume of three representative methods: AsyncDiff \cite{Asyncdiff} with stage-wise communication, Ring Attention \cite{ringattention} used in xDiT \cite{xDiT} which performs layer-wise communication, and our proposed ParaStep which operates at the step level. For clarity, we assume that the intermediate feature tensors remain of constant size \(M\) across all layers of the noise predictor, which is typically implemented using DiT or U-Net. We also assume the model has \(L\) attention layers and the parallelism degree is \(p\).


\paragraph{Ring Attention} In each layer, every GPU must gather key and value tensors from all other GPUs. The size of the keys/values from other devices is \(\frac{p-1}{p}M\), and the total communication per GPU per layer is therefore \(2\frac{p-1}{p}M\). Across all layers, the total communication $\mathcal{C}_{\text{Ring}}$ per step is:
\[
\mathcal{C}_{\text{Ring}} = 2L(p - 1)M.
\]

\paragraph{AsyncDiff} Each GPU handles a pipeline stage, performing a broadcast of \((p - 1)M\) data per step. Since this occurs for all \(p\) stages, the total communication $\mathcal{C}_{\text{AsyncDiff}}$ per step is:
\[
\mathcal{C}_{\text{AsyncDiff}} = p(p - 1)M.
\]

\paragraph{ParaStep} Each cycle (of length \(p\)) involves \(p - 1\) send-receive operations and one broadcast. The total communication volume per cycle is $(p-1)M + (p-1)M = 2(p - 1)M$. Since one cycle covers \(p\) denoising steps, the average communication $\mathcal{C}_{\text{ParaStep}}$ per step is:
\[
\mathcal{C}_{\text{ParaStep}} = \frac{2(p - 1)M}{p}.
\]

These results suggest that \textit{ParaStep significantly reduces communication overhead compared to Ring Attention and AsyncDiff, particularly under limited-bandwidth constraints}. 
A detailed derivation of the communication analysis is provided in Appendix~\ref{sec:detailed_communication_analysis}.

\subsection{Leveraging the batching effect for non-compute-intensive models}

\begin{wrapfigure}{r}{0.48\textwidth}  
\vspace{-10pt}
\centering
\includegraphics[width=\linewidth]{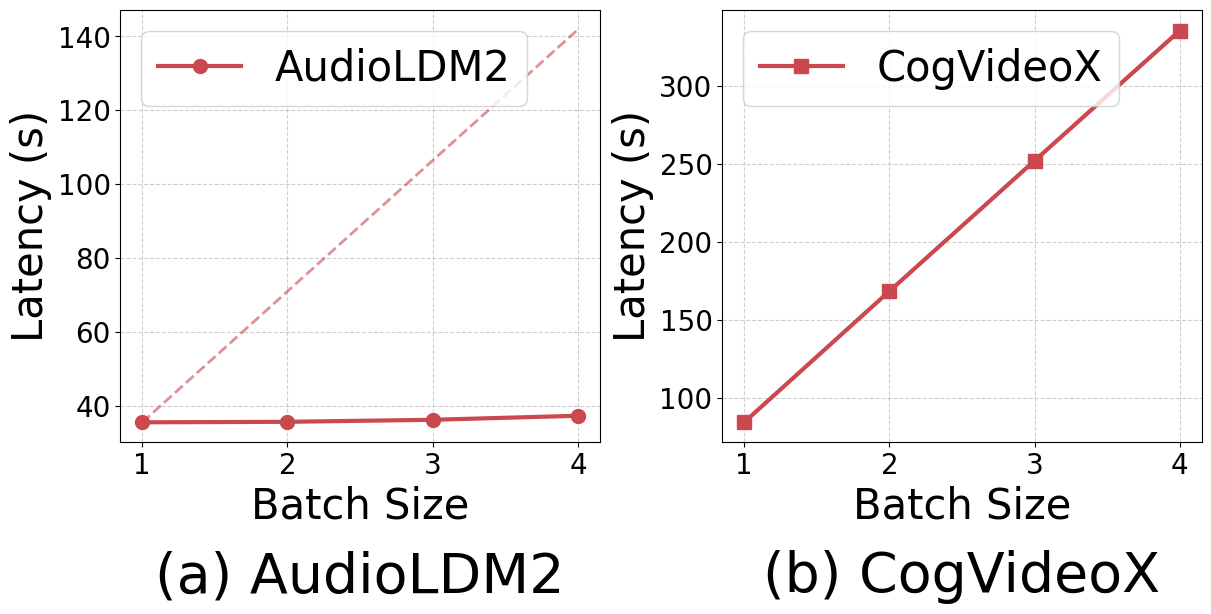}
\vspace{-15pt}
\caption{Batching effect of AudioLDM2-large and CogVideoX-2b.}
\label{fig:batching_effect}
\end{wrapfigure}

Image and video diffusion models are typically computationally intensive, where increasing the batch size leads to a linear increase in latency. In contrast, audio diffusion models exhibit a more favorable batching effect. As shown in Figure~\ref{fig:batching_effect}, increasing the batch size results in only a marginal increase in latency. By leveraging this property, we can transform the parallel execution of the noise predictor into a batched inference process on a single device, as illustrated in Figure~\ref{fig:local_batching}. We refer to this single-device variant as \textbf{BatchStep}. The degree of parallelism $p$ in ParaStep corresponds to the cycle length $s$ in BatchStep. This approach achieves a theoretical reduction in per-step computation of approximately $\frac{s - 1}{s}$, while completely eliminating the need for multi-device parallelism.

\subsection{Dynamic degree of parallelism via TeaCache}

TeaCache \cite{TeaCache} introduces a selective mechanism for reusing previously generated noise, using a larger stride of reuse in less critical timesteps to improve efficiency. We extend ParaStep by integrating the computation schedule of TeaCache. Specifically, the degree of parallelism at each timestep is determined based on the TeaCache schedule. Since ParaStep has low communication overhead, it can achieve similar speedups to TeaCache. Moreover, due to the reuse-then-predict mechanism, our method can offer superior generation quality.

\section{Experiments}

\subsection{Settings}
\label{subsec:settings}
\paragraph{Base models and compared methods}
Diffusion model is a special case of Flow Matching \cite{flow}, where both perform a step-by-step computation process to reconstruct the target data distribution. To demonstrate the versatility of our method, we evaluate it on both diffusion models and Flow Matching model. For the Flow Matching model, we adopt Stable Diffusion 3 (SD3) \cite{sd3} as a representative text-to-image model. For diffusion models, we evaluate on CogVideoX-2b \cite{Cogvideox} and Latte \cite{Latte}, which are text-to-video models, as well as SVD \cite{SVD}, an image-to-video model. We also include AudioLDM2-large \cite{Audioldm2} for audio generation. For comparison, we include three state-of-the-art parallel methods: AsyncDiff \cite{Asyncdiff}, PipeFusion \cite{pipefusion}, and Ring Attention \cite{ringattention}. The number of warm-up steps for ParaStep is set to 1 for AudioLDM2-large, 5 for SD3, 5 for SVD, 13 for CogVideoX-2b, and 18 for Latte. The number of inference steps is set to 200 for AudioLDM2-large and 50 for all other models.


\paragraph{Dataset and evaluation metrics}
We use the MS-COCO 2017 \cite{coco} validation set for text-to-image model, VBench \cite{Vbench} for text-to-video and image-to-video models, and AudioCaps \cite{audiocaps} for audio model, AudioLDM2-large. For text-to-image model, we evaluate performance using Peak Signal-to-Noise Ratio (PSNR), Learned Perceptual Image Patch Similarity (LPIPS) \cite{lpips}, Structural Similarity Index Measure (SSIM), and Fréchet Inception Distance (FID) \cite{fid}. For text-to-video and image-to-video models, we use PSNR, LPIPS, SSIM, and VBench (t2v, i2v) score. For audio model, we utilize Fréchet Audio Distance (FAD) \cite{FAD}, which measures the quality of generated audio.

\paragraph{Hardware}
All experiments are conducted on a machine equipped with 8 NVIDIA 4090 GPUs (24GB each) \cite{nvidia4090}, connected via PCIe Gen3.

\subsection{Quantitative comparison with baselines}

xDiT implements PipeFusion \cite{pipefusion} (xDiT-Pipe) and Ring Attention \cite{ringattention} (xDiT-Ring) to accelerate the denoising process. We compare ParaStep with xDiT-Pipe on the image model SD3. Since xDiT does not currently support PipeFusion for video generation models, we instead compare ParaStep with xDiT-Ring for video generation.
AsyncDiff is another recent parallel method, and we evaluate ParaStep against AsyncDiff on SVD and SD3. Both xDiT-Pipe, AsyncDiff, and ParaStep rely on approximate methods for parallelization, which introduce varying degrees of deviation from the original model. Although xDiT-Ring is theoretically lossless, its computational logic differs from that of the original model, potentially leading to minor discrepancies in generated outputs. As shown in Table~\ref{tab:table-image} and Table~\ref{tab:comparison_baselines}, our method achieves state-of-the-art performance compared to baseline methods on both image and video generation models. For SD3, ParaStep demonstrates the highest speedup while maintaining superior generation quality compared to AsyncDiff and xDiT-Pipe. With a parallelism degree of 2, ParaStep achieves a 1.68$\times$ speedup over the original denoising process, with a FID of only 5.01. For video generation models, ParaStep also achieves the highest speedup among all baseline methods while maintaining the best generation quality across different degrees of parallelism on SVD and CogVideoX-2b. These results confirm that our method is both highly efficient and introduces minimal impact on generation quality.  It is worth noting that, due to the high communication overhead of xDiT-Ring, it provides little to no speedup in our experimental setup.

\begin{table*}[t]
\vspace{-10pt}
\centering
\small
\begin{tabular}{l|cc|cccc}
\toprule
\multirow{2}{*}{\textbf{Method}} & \multicolumn{2}{c|}{\textbf{Efficiency}} & \multicolumn{4}{c}{\textbf{Visual Quality}} \\
 & Speedup ↑ & Latency (s) ↓ & FID ↓ & LPIPS ↓ & PSNR ↑ & SSIM ↑ \\
\midrule
SD3 ($T=50$)                    & 1 & 18.75 & - & - & - & - \\
AsyncDiff ($p=2$)         & 1.61 & 11.62 & 7.11 & 0.2141 & 16.47 & 0.7290 \\
xDiT-Pipe ($p=2$)         & 1.50 & 12.50 & 6.20 & 0.1943 & 16.68 & 0.7498 \\
\textbf{ParaStep} ($p=2$)         & \textbf{1.68} & \textbf{11.16} & \textbf{5.01} & \textbf{0.1362} & \textbf{18.61} & \textbf{0.8157}\\
\bottomrule
\end{tabular}
\caption{Speedup and generation quality on image model SD3, with a resolution of 1440$\times$1440. T is the number of inference steps, and p is the degree of parallelism.}
\label{tab:table-image}
\end{table*}

\begin{table*}[t]
\centering
\small
\begin{tabular}{l|cc|cccc}
\toprule
\multirow{2}{*}{\textbf{Method}} & \multicolumn{2}{c|}{\textbf{Efficiency}} & \multicolumn{4}{c}{\textbf{Visual Quality}} \\
 & Speedup ↑ & Latency (s) ↓ & VBench ↑ & LPIPS ↓ & PSNR ↑ & SSIM ↑ \\
\midrule

\multicolumn{7}{c}{\textbf{SVD (14 frames, 1024×576)}} \\
\midrule
SVD ($T=50$)                    & 1 & 51.04 & \textbf{42.00} & - & - & - \\
AsyncDiff ($p=2$)           & 1.37 & 37.36 & 41.75 & 0.0790 & 25.99 & 0.8366 \\
AsyncDiff ($p=4$)           & 1.80 & 28.43 & 41.56 & 0.1285 & 23.13 & 0.7599 \\
\textbf{ParaStep} ($p=2$)         & 1.69 & 30.27 & 41.90 & \textbf{0.0278} & \textbf{33.35} & \textbf{0.9347} \\
\textbf{ParaStep} ($p=4$)         & \textbf{2.49} & \textbf{20.49} & 41.74 & 0.0732 & 26.99 & 0.8433 \\
\midrule

\multicolumn{7}{c}{\textbf{Latte (16 frames, 512×512)}} \\
\midrule
Latte ($T=50$)                  & 1 & 32.56 & 73.68 & - & - & - \\
xDiT-Ring ($p=2$)           & 1.01 & 32.18 & \textbf{73.87} & \textbf{0.0424} & 32.79 & \textbf{0.9273} \\
xDiT-Ring ($p=4$)           & 1.07 & 30.50 & 73.81 & 0.0431 & \textbf{32.80} & 0.9266 \\
\textbf{ParaStep} ($p=2$)         & 1.43 & 22.80 & 73.76 & 0.0432 & 32.51 & 0.9258 \\
\textbf{ParaStep} ($p=4$)         & \textbf{1.82} & \textbf{17.91} & 73.76 & 0.0533 & 31.10 & 0.9115 \\
\midrule


\multicolumn{7}{c}{\textbf{CogVideoX-2b (45 frames, 512×720)}} \\
\midrule
CogVideoX ($T=50$)              & 1 & 91.89 & 76.95 & - & - & - \\
xDiT-Ring ($p=2$)           & 0.86 & 106.68 & 76.79 & 0.0570 & 32.24 & 0.9284 \\
xDiT-Ring ($p=4$)           & 0.98 & 93.44 & 76.66 & 0.0817 & 29.53 & 0.9028 \\
\textbf{ParaStep} ($p=2$)         & 1.47 & 62.50 & \textbf{76.97} & \textbf{0.0213} & \textbf{37.57} & \textbf{0.9642} \\
\textbf{ParaStep} ($p=4$)         & \textbf{1.93} & \textbf{47.66} & 76.74 & 0.0359 & 34.34 & 0.9505 \\
\bottomrule
\end{tabular}
\caption{Comparison of speedup and generation quality across four video generation models using different parallel methods.}
\label{tab:comparison_baselines}
\end{table*}




\begin{figure}[t]
  \vspace{-15pt}
  \centering
  \begin{minipage}[t]{0.46\linewidth}
    \vspace{0pt}  
    \centering
    \small
    \begin{tabular}{l|cc}
      \toprule
      \textbf{Method} & Latency (s) ↓ & FAD ↓ \\
      \midrule
      AudioLDM2 ($T = 200$)         & 34.80 & 1.6653 \\
      ParaStep ($p = 2$)            & 18.86 & \textbf{1.6651} \\
      ParaStep ($p = 4$)            & 9.86 & 1.6716 \\
      BatchStep ($s = 2$)           & 17.74 & 1.6671     \\
      BatchStep ($s = 4$)           & \textbf{9.45} & 1.6699 \\
      \bottomrule
    \end{tabular}
    \captionof{table}{Generation latency and FAD on AudioLDM2-large. $p$ is the degree of parallelism, $s$ is the cycle length in BatchStep.}
    \label{tab:table-audio}
  \end{minipage}
  \hfill
  \begin{minipage}[t]{0.50\linewidth}
    \vspace{-5pt}  
    \centering
    \includegraphics[width=\linewidth]{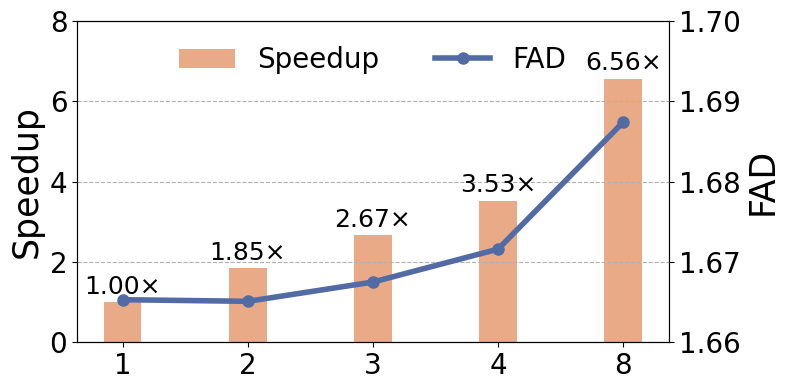}
    \vspace{-15pt}
    \caption{FAD under varying degrees of parallelism using ParaStep. Model: AudioLDM2-large.}
    \label{fig:scalibility}
  \end{minipage}
\end{figure}


\begin{figure}[t]
\vspace{-5pt}
\centering
\begin{minipage}[t]{0.46\linewidth}
\vspace{0pt}  
\centering
\small
\begin{tabular}{l|cc}
\toprule
\textbf{Method}        & Latency (s) ↓ & PSNR ↑ \\
\midrule
TeaCache-slow & 52.25       & 32.25      \\
TeaCache-fast & \textbf{36.19}       & 24.42      \\
\textbf{ParaStep-slow} & 52.61       & \textbf{35.76}      \\
\textbf{ParaStep-fast} & 37.27       & 25.59      \\ 
\bottomrule
\end{tabular}
\captionof{table}{Extend ParaStep by integrating the computation schedule of TeaCache to achieve superior generation quality.}
\label{tab:table-teacache}
\end{minipage}
\hfill
\begin{minipage}[t]{0.50\linewidth}
\vspace{-5pt}  
\centering
\includegraphics[width=\linewidth]{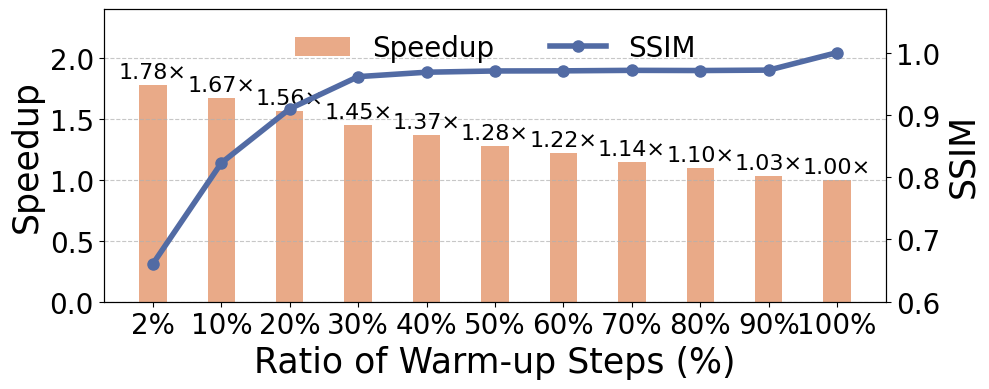}
\vspace{-15pt}
\caption{Impact of the number of warm-up steps on the speedup and generation quality of ParaStep. Degree of parallelism: 2. Model: CogVideoX-2b.}
\label{fig:impact_of_warm}
\end{minipage}
\end{figure}

\subsection{Effect of BatchStep}
We compare the generation latency and FAD of ParaStep and BatchStep on AudioLDM2-large, with the number of warm-up steps set to 1. As shown in Table~\ref{tab:table-audio}, both ParaStep and BatchStep achieve acceleration with only a minor degradation in generation quality. Notably, since BatchStep transforms the parallel execution of the noise predictor into a batched inference process on a single device, it eliminates the need for inter-device communication. As a result, BatchStep achieves greater acceleration than ParaStep.

\begin{wraptable}{r}{0.52\textwidth}
\vspace{-10pt}
\centering
\small
\setlength{\tabcolsep}{4pt}  
\renewcommand{\arraystretch}{1.1}  
\begin{tabular}{l|ccc}
\toprule
\textbf{Method} & \textbf{Total (s)} & \textbf{Comp. (s)} & \textbf{Comm. (s)} \\
\midrule
SVD            & 51.04 & 51.04 & - \\
AsyncDiff ($p=2$) & 39.30 & 30.71 & 8.59 \\
ParaStep ($p=2$)  & 30.52 & 30.42 & 0.10 \\
AsyncDiff ($p=4$) & 31.30 & 21.01 & 10.29 \\
\textbf{ParaStep} ($p=4$)  & \textbf{20.08} & 19.94 & 0.14 \\
\bottomrule
\end{tabular}
\vspace{-5pt}
\caption{Comparison of total latency and breakdown into computation and communication times across different parallel settings.}
\label{tab:latency_breakdown}
\vspace{-10pt}
\end{wraptable}

\subsection{Performance breakdown}
To demonstrate that the observed speedup achieved by ParaStep primarily stems from improved communication efficiency, we conducted performance breakdown on the SVD model. Specifically, we measured both the total latency and the communication-related latency for AsyncDiff and our proposed ParaStep. As shown in the Table \ref{tab:latency_breakdown}, ParaStep exhibits significantly lower communication latency compared to AsyncDiff, which directly contributes to its higher overall speedup.

\subsection{Ablation studies}
\label{subsec:Ablation}
\paragraph{Scalability of our method}
We evaluate the scalability of ParaStep on AudioLDM2-large with a warm-up step count of 1. As shown in Figure~\ref{fig:scalibility}, ParaStep achieves a 6.56× speedup at a parallelism degree of 8, with only a 0.02 increase in FAD, demonstrating that our method can effectively scale to high degrees of parallelism on AudioLDM2-large. Additional scalability evaluations on other models are provided in Appendix~\ref{subsec:scalability_more}.
\paragraph{Impact of warm-up steps}
The larger the number of warm-up steps, the fewer the number of parallelizable steps. Due to Amdahl's law, the overall speedup of ParaStep is limited by:
$
\frac{1}{m + \frac{1 - m}{p}},
$
where \(m\) denotes the ratio of warm-up steps and \(p\) represents the degree of parallelism. 

However, using too few warm-up steps leads to a degradation in generation quality, as the initial denoising steps are more sensitive to prediction errors. As shown in Figure~\ref{fig:impact_of_warm}, setting the warm-up ratio to 30\% offers a favorable trade-off, achieving strong generation performance with only a moderate reduction in speedup. Increasing the ratio from 30\% to 100\% yields only marginal improvements in generation quality, but causes a substantial drop in acceleration.

\paragraph{Extending ParaStep with TeaCache}
By integrating the computation schedule of TeaCache into ParaStep, our method achieves superior generation quality compared to TeaCache, with only a small overhead from communication, as shown in Table~\ref{tab:table-teacache}. Notably, TeaCache-fast adopts a more aggressive reuse schedule than TeaCache-slow. Similarly, ParaStep-slow refers to the integration of the TeaCache-slow schedule into ParaStep, while ParaStep-fast extends ParaStep with the more aggressive TeaCache-fast schedule.

\section{Conclusion}
\label{sec:conclusion}
In this work, we propose ParaStep, a novel parallelization approach for accelerating diffusion model inference. By leveraging a reuse-then-predict mechanism, ParaStep significantly reduces communication overhead while preserving high generation quality. Our approach is grounded in the observation that adjacent denoising steps often exhibit strong similarity, allowing us to bypass costly computations without substantial performance degradation. We further extend this approach with a single-device variant, BatchStep, which transforms parallel execution into efficient batched inference for non-compute-intensive models, such as audio diffusion models. Extensive experimental evaluations confirm that ParaStep delivers superior speedup across diverse modalities, including image, video, and audio generation, while maintaining competitive quality metrics. 



\section{Acknowledgments}
This work is sponsored by the National Natural Science Foundation of China (62472273, 62232015).

\clearpage %
{
    \small
    \bibliographystyle{plain}
    \bibliography{main}

\begin{thebibliography}{10}

\bibitem{approximate}
Shubham Agarwal, Subrata Mitra, Sarthak Chakraborty, Srikrishna Karanam, Koyel Mukherjee, and Shiv~Kumar Saini.
\newblock Approximate caching for efficiently serving $\{$Text-to-Image$\}$ diffusion models.
\newblock In {\em 21st USENIX Symposium on Networked Systems Design and Implementation (NSDI 24)}, pages 1173--1189, 2024.

\bibitem{SVD}
Andreas Blattmann, Tim Dockhorn, Sumith Kulal, Daniel Mendelevitch, Maciej Kilian, Dominik Lorenz, Yam Levi, Zion English, Vikram Voleti, Adam Letts, et~al.
\newblock Stable video diffusion: Scaling latent video diffusion models to large datasets.
\newblock {\em arXiv preprint arXiv:2311.15127}, 2023.

\bibitem{Asyncdiff}
Zigeng Chen, Xinyin Ma, Gongfan Fang, Zhenxiong Tan, and Xinchao Wang.
\newblock Asyncdiff: Parallelizing diffusion models by asynchronous denoising.
\newblock {\em arXiv preprint arXiv:2406.06911}, 2024.

\bibitem{sd3}
Patrick Esser, Sumith Kulal, Andreas Blattmann, Rahim Entezari, Jonas M{\"u}ller, Harry Saini, Yam Levi, Dominik Lorenz, Axel Sauer, Frederic Boesel, et~al.
\newblock Scaling rectified flow transformers for high-resolution image synthesis.
\newblock In {\em Forty-first international conference on machine learning}, 2024.

\bibitem{xDiT}
Jiarui Fang, Jinzhe Pan, Xibo Sun, Aoyu Li, and Jiannan Wang.
\newblock xdit: an inference engine for diffusion transformers (dits) with massive parallelism.
\newblock {\em arXiv preprint arXiv:2411.01738}, 2024.

\bibitem{pipefusion}
Jiarui Fang, Jinzhe Pan, Jiannan Wang, Aoyu Li, and Xibo Sun.
\newblock Pipefusion: Patch-level pipeline parallelism for diffusion transformers inference.
\newblock {\em arXiv preprint arXiv:2405.14430}, 2024.

\bibitem{gao2025diffusion}
Ruiqi Gao, Emiel Hoogeboom, Jonathan Heek, Valentin~De Bortoli, Kevin~Patrick Murphy, and Tim Salimans.
\newblock Diffusion models and gaussian flow matching: Two sides of the same coin.
\newblock In {\em The Fourth Blogpost Track at ICLR 2025}, 2025.

\bibitem{audio3}
Zhifang Guo, Jianguo Mao, Rui Tao, Long Yan, Kazushige Ouchi, Hong Liu, and Xiangdong Wang.
\newblock Audio generation with multiple conditional diffusion model.
\newblock In {\em Proceedings of the AAAI Conference on Artificial Intelligence}, volume~38, pages 18153--18161, 2024.

\bibitem{fid}
Martin Heusel, Hubert Ramsauer, Thomas Unterthiner, Bernhard Nessler, and Sepp Hochreiter.
\newblock Gans trained by a two time-scale update rule converge to a local nash equilibrium.
\newblock {\em Advances in neural information processing systems}, 30, 2017.

\bibitem{ddpm}
Jonathan Ho, Ajay Jain, and Pieter Abbeel.
\newblock Denoising diffusion probabilistic models.
\newblock {\em Advances in neural information processing systems}, 33:6840--6851, 2020.

\bibitem{audio2}
Rongjie Huang, Jiawei Huang, Dongchao Yang, Yi~Ren, Luping Liu, Mingze Li, Zhenhui Ye, Jinglin Liu, Xiang Yin, and Zhou Zhao.
\newblock Make-an-audio: Text-to-audio generation with prompt-enhanced diffusion models.
\newblock In {\em International Conference on Machine Learning}, pages 13916--13932. PMLR, 2023.

\bibitem{Vbench}
Ziqi Huang, Yinan He, Jiashuo Yu, Fan Zhang, Chenyang Si, Yuming Jiang, Yuanhan Zhang, Tianxing Wu, Qingyang Jin, Nattapol Chanpaisit, et~al.
\newblock Vbench: Comprehensive benchmark suite for video generative models.
\newblock In {\em Proceedings of the IEEE/CVF Conference on Computer Vision and Pattern Recognition}, pages 21807--21818, 2024.

\bibitem{FAD}
Kevin Kilgour, Mauricio Zuluaga, Dominik Roblek, and Matthew Sharifi.
\newblock Fr{\'{e}}chet audio distance: {A} reference-free metric for evaluating music enhancement algorithms.
\newblock In {\em 20th Annual Conference of the International Speech Communication Association, Interspeech 2019, Graz, Austria, September 15-19, 2019}, pages 2350--2354. {ISCA}, 2019.

\bibitem{audiocaps}
Chris~Dongjoo Kim, Byeongchang Kim, Hyunmin Lee, and Gunhee Kim.
\newblock {A}udio{C}aps: Generating captions for audios in the wild.
\newblock In Jill Burstein, Christy Doran, and Thamar Solorio, editors, {\em Proceedings of the 2019 Conference of the North {A}merican Chapter of the Association for Computational Linguistics: Human Language Technologies, Volume 1 (Long and Short Papers)}, pages 119--132, Minneapolis, Minnesota, June 2019. Association for Computational Linguistics.

\bibitem{Hunyuanvideo}
Weijie Kong, Qi~Tian, Zijian Zhang, Rox Min, Zuozhuo Dai, Jin Zhou, Jiangfeng Xiong, Xin Li, Bo~Wu, Jianwei Zhang, et~al.
\newblock Hunyuanvideo: A systematic framework for large video generative models.
\newblock {\em arXiv preprint arXiv:2412.03603}, 2024.

\bibitem{flux2024}
Black~Forest Labs.
\newblock Flux.
\newblock \url{https://github.com/black-forest-labs/flux}, 2024.

\bibitem{distrifusion}
Muyang Li, Tianle Cai, Jiaxin Cao, Qinsheng Zhang, Han Cai, Junjie Bai, Yangqing Jia, Kai Li, and Song Han.
\newblock Distrifusion: Distributed parallel inference for high-resolution diffusion models.
\newblock In {\em Proceedings of the IEEE/CVF Conference on Computer Vision and Pattern Recognition}, pages 7183--7193, 2024.

\bibitem{qdiffusion}
Xiuyu Li, Yijiang Liu, Long Lian, Huanrui Yang, Zhen Dong, Daniel Kang, Shanghang Zhang, and Kurt Keutzer.
\newblock Q-diffusion: Quantizing diffusion models.
\newblock In {\em Proceedings of the IEEE/CVF International Conference on Computer Vision}, pages 17535--17545, 2023.

\bibitem{hunyuandit}
Zhimin Li, Jianwei Zhang, Qin Lin, Jiangfeng Xiong, Yanxin Long, Xinchi Deng, Yingfang Zhang, Xingchao Liu, Minbin Huang, Zedong Xiao, et~al.
\newblock Hunyuan-dit: A powerful multi-resolution diffusion transformer with fine-grained chinese understanding.
\newblock {\em arXiv preprint arXiv:2405.08748}, 2024.

\bibitem{coco}
Tsung-Yi Lin, Michael Maire, Serge Belongie, James Hays, Pietro Perona, Deva Ramanan, Piotr Doll{\'a}r, and C~Lawrence Zitnick.
\newblock Microsoft coco: Common objects in context.
\newblock In {\em Computer vision--ECCV 2014: 13th European conference, zurich, Switzerland, September 6-12, 2014, proceedings, part v 13}, pages 740--755. Springer, 2014.

\bibitem{flow}
Yaron Lipman, Ricky~TQ Chen, Heli Ben-Hamu, Maximilian Nickel, and Matt Le.
\newblock Flow matching for generative modeling.
\newblock {\em arXiv preprint arXiv:2210.02747}, 2022.

\bibitem{TeaCache}
Feng Liu, Shiwei Zhang, Xiaofeng Wang, Yujie Wei, Haonan Qiu, Yuzhong Zhao, Yingya Zhang, Qixiang Ye, and Fang Wan.
\newblock Timestep embedding tells: It's time to cache for video diffusion model.
\newblock {\em arXiv preprint arXiv:2411.19108}, 2024.

\bibitem{ringattention}
Hao Liu, Matei Zaharia, and Pieter Abbeel.
\newblock Ring attention with blockwise transformers for near-infinite context.
\newblock {\em arXiv preprint arXiv:2310.01889}, 2023.

\bibitem{audioldm}
Haohe Liu, Zehua Chen, Yi~Yuan, Xinhao Mei, Xubo Liu, Danilo Mandic, Wenwu Wang, and Mark~D Plumbley.
\newblock Audioldm: text-to-audio generation with latent diffusion models.
\newblock In {\em Proceedings of the 40th International Conference on Machine Learning}, pages 21450--21474, 2023.

\bibitem{Audioldm2}
Haohe Liu, Yi~Yuan, Xubo Liu, Xinhao Mei, Qiuqiang Kong, Qiao Tian, Yuping Wang, Wenwu Wang, Yuxuan Wang, and Mark~D Plumbley.
\newblock Audioldm 2: Learning holistic audio generation with self-supervised pretraining.
\newblock {\em IEEE/ACM Transactions on Audio, Speech, and Language Processing}, 2024.

\bibitem{Latte}
Xin Ma, Yaohui Wang, Gengyun Jia, Xinyuan Chen, Ziwei Liu, Yuan-Fang Li, Cunjian Chen, and Yu~Qiao.
\newblock Latte: Latent diffusion transformer for video generation.
\newblock {\em arXiv preprint arXiv:2401.03048}, 2024.

\bibitem{meng2023distillation}
Chenlin Meng, Robin Rombach, Ruiqi Gao, Diederik Kingma, Stefano Ermon, Jonathan Ho, and Tim Salimans.
\newblock On distillation of guided diffusion models.
\newblock In {\em Proceedings of the IEEE/CVF Conference on Computer Vision and Pattern Recognition}, pages 14297--14306, 2023.

\bibitem{nvidia4090}
{NVIDIA Corporation}.
\newblock {NVIDIA GeForce RTX 4090 GPU}, 2022.

\bibitem{dit}
William Peebles and Saining Xie.
\newblock Scalable diffusion models with transformers.
\newblock In {\em Proceedings of the IEEE/CVF international conference on computer vision}, pages 4195--4205, 2023.

\bibitem{image1}
Aditya Ramesh, Prafulla Dhariwal, Alex Nichol, Casey Chu, and Mark Chen.
\newblock Hierarchical text-conditional image generation with clip latents.
\newblock {\em arXiv preprint arXiv:2204.06125}, 1(2):3, 2022.

\bibitem{image2}
Robin Rombach, Andreas Blattmann, Dominik Lorenz, Patrick Esser, and Bj{\"o}rn Ommer.
\newblock High-resolution image synthesis with latent diffusion models.
\newblock In {\em Proceedings of the IEEE/CVF conference on computer vision and pattern recognition}, pages 10684--10695, 2022.

\bibitem{unet}
Olaf Ronneberger, Philipp Fischer, and Thomas Brox.
\newblock U-net: Convolutional networks for biomedical image segmentation.
\newblock In {\em Medical image computing and computer-assisted intervention--MICCAI 2015: 18th international conference, Munich, Germany, October 5-9, 2015, proceedings, part III 18}, pages 234--241. Springer, 2015.

\bibitem{image3}
Chitwan Saharia, William Chan, Saurabh Saxena, Lala Li, Jay Whang, Emily~L Denton, Kamyar Ghasemipour, Raphael Gontijo~Lopes, Burcu Karagol~Ayan, Tim Salimans, et~al.
\newblock Photorealistic text-to-image diffusion models with deep language understanding.
\newblock {\em Advances in neural information processing systems}, 35:36479--36494, 2022.

\bibitem{sauer2024distillation}
Axel Sauer, Dominik Lorenz, Andreas Blattmann, and Robin Rombach.
\newblock Adversarial diffusion distillation.
\newblock In {\em European Conference on Computer Vision}, pages 87--103. Springer, 2024.

\bibitem{audio1}
Flavio Schneider.
\newblock Archisound: Audio generation with diffusion.
\newblock {\em arXiv preprint arXiv:2301.13267}, 2023.

\bibitem{shang2023post}
Yuzhang Shang, Zhihang Yuan, Bin Xie, Bingzhe Wu, and Yan Yan.
\newblock Post-training quantization on diffusion models.
\newblock In {\em Proceedings of the IEEE/CVF conference on computer vision and pattern recognition}, pages 1972--1981, 2023.

\bibitem{deepcache}
Mengwei Xu, Mengze Zhu, Yunxin Liu, Felix~Xiaozhu Lin, and Xuanzhe Liu.
\newblock Deepcache: Principled cache for mobile deep vision.
\newblock In {\em Proceedings of the 24th annual international conference on mobile computing and networking}, pages 129--144, 2018.

\bibitem{Cogvideox}
Zhuoyi Yang, Jiayan Teng, Wendi Zheng, Ming Ding, Shiyu Huang, Jiazheng Xu, Yuanming Yang, Wenyi Hong, Xiaohan Zhang, Guanyu Feng, et~al.
\newblock Cogvideox: Text-to-video diffusion models with an expert transformer.
\newblock {\em arXiv preprint arXiv:2408.06072}, 2024.

\bibitem{morecite1}
Hancheng Ye, Jiakang Yuan, Renqiu Xia, Xiangchao Yan, Tao Chen, Junchi Yan, Botian Shi, and Bo~Zhang.
\newblock Training-free adaptive diffusion with bounded difference approximation strategy.
\newblock {\em Advances in Neural Information Processing Systems}, 37:306--332, 2024.

\bibitem{consisid}
Shenghai Yuan, Jinfa Huang, Xianyi He, Yunyuan Ge, Yujun Shi, Liuhan Chen, Jiebo Luo, and Li~Yuan.
\newblock Identity-preserving text-to-video generation by frequency decomposition.
\newblock {\em arXiv preprint arXiv:2411.17440}, 2024.

\bibitem{lpips}
Richard Zhang, Phillip Isola, Alexei~A Efros, Eli Shechtman, and Oliver Wang.
\newblock The unreasonable effectiveness of deep features as a perceptual metric.
\newblock In {\em Proceedings of the IEEE conference on computer vision and pattern recognition}, pages 586--595, 2018.

\bibitem{morecite2}
Chang Zou, Xuyang Liu, Ting Liu, Siteng Huang, and Linfeng Zhang.
\newblock Accelerating diffusion transformers with token-wise feature caching.
\newblock {\em arXiv preprint arXiv:2410.05317}, 2024.

\end{thebibliography}
}

\newpage
\appendix

\section*{\centering Appendix}

\section{Round-based implementation of ParaStep}
\label{sec:Implementation}

\begin{algorithm}
\caption{\textbf{Round-based implementation of ParaStep}}
\label{alg:algorithm_ParaStep}
\begin{algorithmic}[1]
\Require Number of inference steps \(T\), degree of parallelism \(p\), rank of GPU \(rank\)
\Ensure Final output \(\mathbf{x}_0\)
\State Initialize noisy sample: \(\mathbf{x}_T \sim \mathcal{N}(\mathbf{0}, \mathbf{I})\)
\State Initialize round counter: \(\textit{round} \gets 0\)

\For{\(t = T, T-1, \dots, 1\)}
    \If{$t \in \text{warmup\_steps}$}
        \Comment{Perform the original computation process during warm-up steps}
        \State Compute noise: \(\epsilon_t = \epsilon_{\theta}(\mathbf{x}_t, t)\)
        \State Cache noise: \(\epsilon_{\text{cache}} \gets \epsilon_t\)
        \State Update sample: \(\mathbf{x}_{t-1} = \text{Scheduler}(\mathbf{x}_t, t, \epsilon_t)\)
    \Else
        \Comment{Perform the ParaStep computation process in non-warm-up steps}
        \If{\(rank = \textit{round}\)}
            \State Compute noise: \(\epsilon_t = \epsilon_{\theta}(\mathbf{x}_t, t)\)
            \State Cache noise: \(\epsilon_{\text{cache}} \gets \epsilon_t\)
            \If{\(rank \neq 0\)}
                \State \textbf{Send} \(\epsilon_t\) to GPU 0
            \EndIf
        \Else
            \State Reuse cached noise: \(\epsilon_t \gets \epsilon_{\text{cache}}\)
            \If{\(rank = 0\)}
                \State \textbf{Receive} \(\epsilon_t\) from GPU \(\textit{round}\)
            \EndIf
        \EndIf
        \State Update sample: \(\mathbf{x}_{t-1} = \text{Scheduler}(\mathbf{x}_t, t, \epsilon_t)\)
        \If{\(\textit{round} = p - 1\)}
            \State \textbf{Broadcast} \(\mathbf{x}_{t-1}\) from GPU 0 to all GPUs
        \EndIf
        \State Update round counter: \(\textit{round} \gets (\textit{round} + 1) \mod p\)
    \EndIf
\EndFor

\State \Return \(\mathbf{x}_0\)

\end{algorithmic}
\end{algorithm}


We formalize ParaStep with a round-based algorithm, presented in Algorithm~\ref{alg:algorithm_ParaStep}. Each GPU is assigned a unique identifier \texttt{rank}, and a variable \texttt{round} designates the current master GPU. Assuming a parallelism degree of $p$, ParaStep operates in cycles of length $p$. The master GPU is the one whose \texttt{rank} matches the current \texttt{round}, and the root GPU is designated as \texttt{rank} = 0.

In each round, the master GPU invokes the noise predictor $\epsilon_{\theta}$ to estimate $\epsilon_t$, while all other GPUs skip this computation. The predicted noise is then sent to the root GPU, which uses it to compute the corresponding noisy sample via the scheduler. In the final round of each cycle (i.e., \texttt{round} = $p-1$), the root GPU obtains the last noisy sample and broadcasts it to all GPUs.

Each GPU performs one forward pass of the noise predictor per cycle. Since scheduler operations are negligible in cost, noise prediction across all devices is effectively parallelized. Consequently, the total latency of the denoising process is reduced by a factor of $p$, achieving a theoretical speedup of $p$.

\section{Detailed derivation of communication analysis}
\label{sec:detailed_communication_analysis}

In this section, we provide a more detailed derivation of the communication volumes for the three parallelism strategies discussed in Section~\ref{subsec:communitation}: \textbf{Ring Attention}, \textbf{AsyncDiff}, and \textbf{ParaStep}. We assume that the intermediate feature tensors remain of constant size \(M\) across all layers of the noise predictor, and that the model contains \(L\) attention layers. The parallelism degree is denoted as \(p\).

\subsection{Ring Attention}

In Ring Attention \cite{ringattention}, the keys and values on each GPU are split into \(p\) partitions, each of size $\frac{1}{p}M$. Before each attention operation, GPUs must gather the keys and values from all other GPUs in a ring-style communication pattern to obtain the complete set of keys and values. 

For each GPU, this process involves:

Gathering $p-1$ partitions, each of size \(\frac{1}{p}M\), for both keys and values.
- The total communication volume per GPU per layer is therefore:
\[
\frac{p-1}{p}M \text{ (keys)} + \frac{p-1}{p}M \text{ (values)} = \frac{2(p-1)}{p}M.
\]

Since this communication occurs in every attention layer, the total communication volume per GPU per step is:
\[
\mathcal{C}_{\text{Ring, per GPU}} = \frac{2(p-1)}{p}M \cdot L.
\]

With \(p\) GPUs, the total communication volume across all GPUs per step is:
\[
\mathcal{C}_{\text{Ring}} = 2L(p - 1)M.
\]

\subsection{AsyncDiff}

In AsyncDiff \cite{Asyncdiff}, the model is divided into \(p\) pipeline stages, each assigned to a separate GPU. In each step, every stage (or GPU) must broadcast its outputs to all other GPUs, resulting in a communication volume of:
\[
(p-1)M \text{ (per GPU per step)}.
\]

Since the model contains \(p\) stages, the total communication volume per step across all GPUs is:
\[
\mathcal{C}_{\text{AsyncDiff}} = p(p - 1)M.
\]

\subsection{ParaStep}

ParaStep adopts a round-based communication pattern, where each cycle contains \(p\) steps. In each cycle:

\begin{enumerate}
    \item \textbf{Noise transfer}: Each GPU (except for GPU 0) sends its predicted noise \(\epsilon\) to GPU 0, resulting in:
    \[
    (p - 1)M \text{ (noise transfer per cycle)}.
    \]

    \item \textbf{Sample broadcast}: In the final step of each cycle, GPU 0 broadcasts the predicted noisy sample \(\mathbf{x}\) to all other GPUs, leading to:
    \[
    (p - 1)M \text{ (sample broadcast per cycle)}.
    \]
\end{enumerate}
The total communication volume per cycle is therefore:
\[
2(p - 1)M.
\]

Since one cycle covers \(p\) denoising steps, the average communication volume per step is:
\[
\mathcal{C}_{\text{ParaStep}} = \frac{2(p - 1)M}{p}.
\]

\section{Limitations}
\label{sec:limitations}
The main limitations of our method are as follows:

Firstly, ParaStep is a parallelization method, which requires additional computational resources to achieve speedup, effectively trading compute for efficiency.

Secondly, ParaStep replicates the entire noise predictor across all GPUs, which means it cannot reduce the per-GPU memory consumption of the noise predictor. A typical diffusion pipeline consists of several components, including a text encoder, a noise predictor (e.g., DiT or U-Net), and a VAE, among others. Among these, the text encoder is often the primary memory bottleneck. Unlike the noise predictor, the text encoder is not replicated; instead, it is partitioned into multiple stages distributed across devices, thereby reducing memory usage on each GPU.

\section{Societal impacts}
\label{sec:impacts}
We propose ParaStep, a novel parallelization approach for accelerating diffusion model inference. ParaStep achieves superior generation quality compared to state-of-the-art parallelization methods, while delivering greater speedup through lightweight, step-wise communication.

ParaStep can be applied in commercial settings to accelerate compute-intensive diffusion models, such as vision-based models, thereby making diffusion technologies more accessible to users and researchers without access to expensive, data center–scale infrastructure.
For non-compute-intensive diffusion models, our proposed variant BatchStep enables speedup on a single device, offering a nearly free performance gain with minimal hardware cost.

\section{Supplementary experiments}

\subsection{Using ParaStep to accelerate HunyuanVideo}
We compare the speedup and generation quality of ParaStep and xDiT-Ring on HunyuanVideo. Since HunyuanVideo requires more GPU memory than the 24GB capacity of an NVIDIA 4090, we apply quantization to reduce memory consumption. In this experiment, each video contains 5 frames with a resolution of 180×180.

As shown in Table~\ref{tab:hunyuanvideo_performance}, ParaStep achieves a 2.67× speedup compared to the original model, with a PSNR of 30.01, indicating that the generated results are highly similar to those of the original model. Notably, when the degree of parallelism is set to 2, the VBench score of ParaStep is even higher than that of the original model.

Although xDiT-Ring is theoretically lossless, its computational logic differs from that of the original model, potentially leading to minor discrepancies in generated outputs. As a result, the generation quality of xDiT-Ring is generally better than that of ParaStep. However, the significant communication overhead of xDiT-Ring severely limits its speedup, achieving only a 1.08× improvement, which is substantially lower than that of ParaStep.

\begin{table*}[t]
\centering
\small
\begin{tabular}{l|cc|cccc}
\toprule
\textbf{Method} & \textbf{Speedup ↑} & \textbf{Latency (s) ↓} & \textbf{VBench ↑} & \textbf{LPIPS ↓} & \textbf{PSNR ↑} & \textbf{SSIM ↑} \\
\midrule
HunyuanVideo (T=50) & 1 & 27.35 & 72.93 & - & - & - \\
xDiT-Ring ($p=2$) & 1.08 & 25.22 & 72.89 & \textbf{0.0071} & \textbf{37.04} & \textbf{0.9629} \\
xDiT-Ring ($p=4$) & 1.07 & 25.45 & 72.89 & \textbf{0.0071} & \textbf{37.04} & \textbf{0.9629} \\
\textbf{ParaStep} ($p=2$) & 1.79 & 15.28 & \textbf{73.00} & 0.0175 & 34.24 & 0.9482 \\
\textbf{ParaStep} ($p=4$) & \textbf{2.67} & \textbf{10.25} & 72.77 & 0.0468 & 30.11 & 0.9075 \\
\bottomrule
\end{tabular}
\caption{Speedup and generation quality on HunyuanVideo.}
\label{tab:hunyuanvideo_performance}
\end{table*}

\begin{figure}[H]
\centering
\begin{minipage}[t]{0.48\textwidth}
  \centering
  \includegraphics[width=\linewidth]{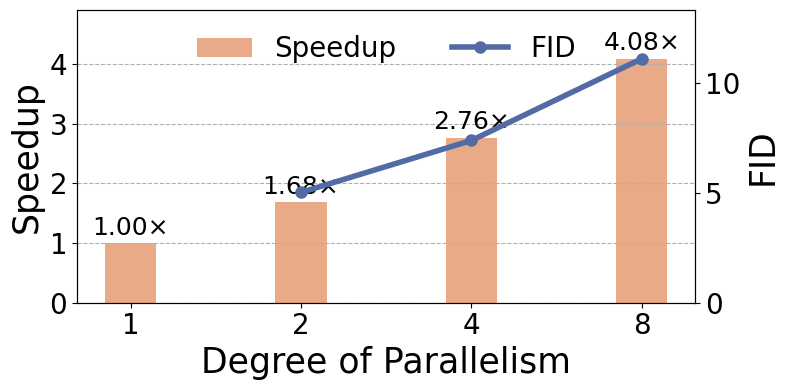}
  \caption{Generation quality under varying degrees of parallelism using ParaStep. Model: SD3.}
  \label{fig:scalibility_sd3}
\end{minipage}%
\hfill
\begin{minipage}[t]{0.48\textwidth}
  \centering
  \includegraphics[width=\linewidth]{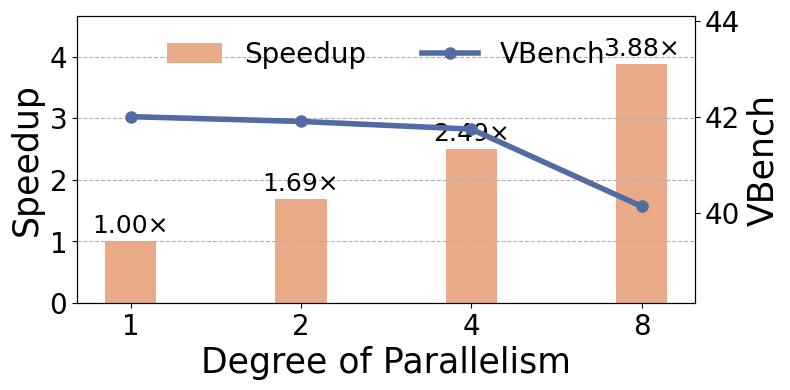}
  \caption{Generation quality under varying degrees of parallelism using ParaStep. Model: SVD.}
  \label{fig:scalibility_svd}
\end{minipage}%
\end{figure}

\begin{figure}[H]
\centering
\begin{minipage}[t]{0.48\textwidth}
  \centering
  \includegraphics[width=\linewidth]{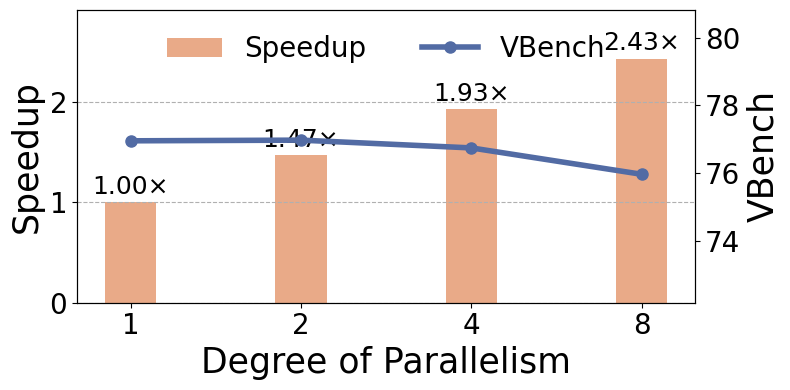}
  \caption{Generation quality under varying degrees of parallelism using ParaStep. Model: CogVideoX-2b.}
  \label{fig:scalibility_cogvideox}
\end{minipage}%
\hfill
\begin{minipage}[t]{0.48\textwidth}
  \centering
  \includegraphics[width=\linewidth]{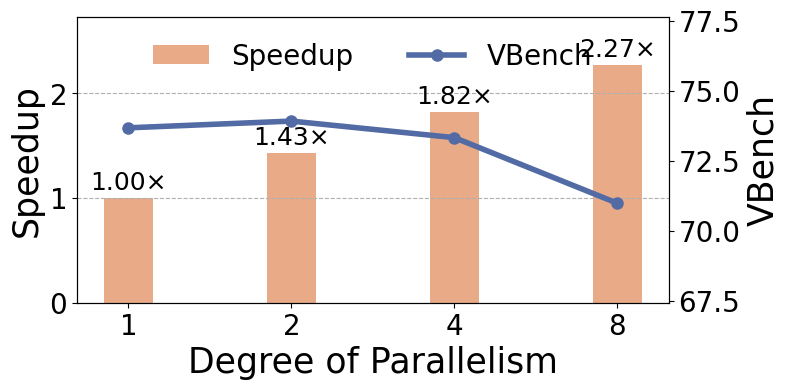}
  \caption{Generation quality under varying degrees of parallelism using ParaStep. Model: Latte.}
  \label{fig:scalibility_latte}
\end{minipage}%
\end{figure}


\subsection{Scalability of ParaStep on SD3, SVD, CogVideoX-2b, and Latte}
\label{subsec:scalability_more}
We evaluate the speedup and generation quality of ParaStep under varying degrees of parallelism on SD3, SVD, CogVideoX-2b, and Latte. The number of warm-up steps for ParaStep is set to 5 for SD3, 5 for SVD, 13 for CogVideoX-2b, and 18 for Latte, with the number of inference steps fixed at 50. As shown in Figure~\ref{fig:scalibility_sd3} and Figure~\ref{fig:scalibility_svd}, with a parallelism degree of 8, ParaStep achieves a 4.08× speedup on SD3 with a FID of only 11.08, and a 3.88× speedup on SVD with a minor VBench score drop of less than 2. However, for CogVideoX-2b and Latte, the speedup is less significant, as shown in Figure~\ref{fig:scalibility_cogvideox} and Figure~\ref{fig:scalibility_latte}, due to the larger number of warm-up steps required for these models.

\subsection{Comparison with single-device efficient inference approach}
Single-device efficiency methods such as DeepCache can reduce the latency of diffusion models without requiring additional computational resources. However, directly caching used in DeepCache may cause a significant degradation in generation quality. In contrast, our ParaStep, based on the Reuse-then-Predict mechanism, offers a balanced trade-off by leveraging additional computational resources to achieve higher speedups with minimal quality loss.

To assess the efficiency of ParaStep, we compare it against both DeepCache and AsyncDiff on the SVD model, which adopts the deterministic \textit{EulerDiscreteScheduler}. We denote the cache interval in DeepCache as $s$, and the degree of parallelism in AsyncDiff and ParaStep as $p$. As shown in Table~\ref{tab:latency_quality_comparison_single_device_efficiency}, ParaStep consistently outperforms both baselines in terms of generation quality and latency.

\begin{table*}[t]
\centering
\small
\setlength{\tabcolsep}{6pt}  
\renewcommand{\arraystretch}{1.05}  
\begin{tabular}{l|c|ccc}
\toprule
\textbf{Method} & \textbf{Latency (s) ↓} & \textbf{LPIPS ↓} & \textbf{PSNR ↑} & \textbf{SSIM ↑} \\
\midrule
SVD ($T=50$)      & 52.23  & -        & -        & -        \\
DeepCache ($s=2$) & 36.42  & 0.0399   & 32.7247  & 0.9155   \\
DeepCache ($s=4$) & 26.90  & 0.0847   & 26.4912  & 0.8415   \\
DeepCache ($s=6$) & 23.75  & 0.1397   & 23.1894  & 0.7728   \\
DeepCache ($s=8$) & 22.24  & 0.1973   & 19.9736  & 0.7160   \\
AsyncDiff ($p=2$) & 39.28  & 0.0861   & 24.0769  & 0.8328   \\
AsyncDiff ($p=4$) & 29.65  & 0.1306   & 21.1599  & 0.7505   \\
\textbf{ParaStep} $(p=2)$ & 30.61  & \textbf{0.0283} & \textbf{32.8624} & \textbf{0.9358} \\
\textbf{ParaStep} $(p=4)$ & \textbf{20.36} & 0.0751 & 24.9949 & 0.8448 \\
\bottomrule
\end{tabular}
\caption{Comparison of latency and generation quality across different caching and parallelization strategies. }
\label{tab:latency_quality_comparison_single_device_efficiency}
\end{table*}



\subsection{Visualization}
\begin{figure}
  \centering
  \includegraphics[width=\linewidth]{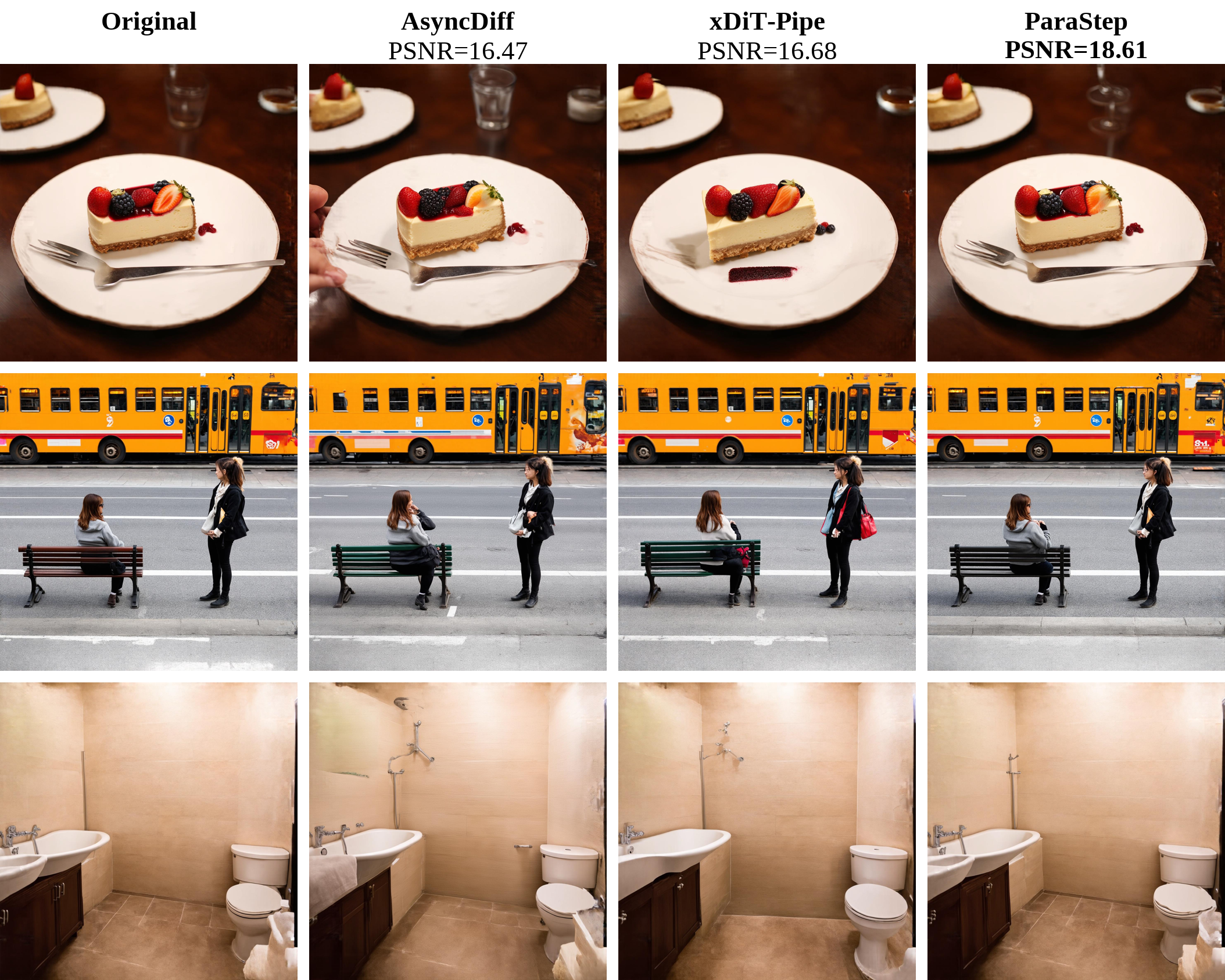}
  \caption{Comparison of generation quality on SD3 using competing methods. PSNR is used to measure the similarity to the original model, with higher values indicating better quality.}
  \label{fig:visualization}
\end{figure}

We visualize the outputs of AsyncDiff, xDiT-Pipe, and ParaStep on SD3. The number of inference steps is set to 50, with a resolution of 1440×1440. As shown in Figure~\ref{fig:visualization}, ParaStep demonstrates superior generation quality compared to AsyncDiff and xDiT-Pipe.

In the first row of Figure~\ref{fig:visualization}, compared to the original diffusion model, AsyncDiff generates an extra hand, while xDiT-Pipe fails to generate a fork and a water cup.
In the second row, the logo in the right corner of the bus generated by AsyncDiff differs from that of the original model, and the bag generated by xDiT-Pipe is red, which is inconsistent with the original output.
In the third row, both AsyncDiff and xDiT-Pipe hallucinate an additional shower that does not appear in the original result.

In contrast, ParaStep achieves higher consistency with the original diffusion model across all examples.

\clearpage





\newpage
\section*{NeurIPS Paper Checklist}

\begin{enumerate}

\item {\bf Claims}
    \item[] Question: Do the main claims made in the abstract and introduction accurately reflect the paper's contributions and scope?
    \item[] Answer: \answerYes{}{} 
    \item[] Justification: The abstract and introduction clearly state the claims made, including the contributions, analysis and important experiments results.
    \item[] Guidelines:
    \begin{itemize}
        \item The answer NA means that the abstract and introduction do not include the claims made in the paper.
        \item The abstract and/or introduction should clearly state the claims made, including the contributions made in the paper and important assumptions and limitations. A No or NA answer to this question will not be perceived well by the reviewers. 
        \item The claims made should match theoretical and experimental results, and reflect how much the results can be expected to generalize to other settings. 
        \item It is fine to include aspirational goals as motivation as long as it is clear that these goals are not attained by the paper. 
    \end{itemize}

\item {\bf Limitations}
    \item[] Question: Does the paper discuss the limitations of the work performed by the authors?
    \item[] Answer: \answerYes{} 
    \item[] Justification: Limitations are discussed in Appendix \ref{sec:limitations}.
    \item[] Guidelines:
    \begin{itemize}
        \item The answer NA means that the paper has no limitation while the answer No means that the paper has limitations, but those are not discussed in the paper. 
        \item The authors are encouraged to create a separate "Limitations" section in their paper.
        \item The paper should point out any strong assumptions and how robust the results are to violations of these assumptions (e.g., independence assumptions, noiseless settings, model well-specification, asymptotic approximations only holding locally). The authors should reflect on how these assumptions might be violated in practice and what the implications would be.
        \item The authors should reflect on the scope of the claims made, e.g., if the approach was only tested on a few datasets or with a few runs. In general, empirical results often depend on implicit assumptions, which should be articulated.
        \item The authors should reflect on the factors that influence the performance of the approach. For example, a facial recognition algorithm may perform poorly when image resolution is low or images are taken in low lighting. Or a speech-to-text system might not be used reliably to provide closed captions for online lectures because it fails to handle technical jargon.
        \item The authors should discuss the computational efficiency of the proposed algorithms and how they scale with dataset size.
        \item If applicable, the authors should discuss possible limitations of their approach to address problems of privacy and fairness.
        \item While the authors might fear that complete honesty about limitations might be used by reviewers as grounds for rejection, a worse outcome might be that reviewers discover limitations that aren't acknowledged in the paper. The authors should use their best judgment and recognize that individual actions in favor of transparency play an important role in developing norms that preserve the integrity of the community. Reviewers will be specifically instructed to not penalize honesty concerning limitations.
    \end{itemize}

\item {\bf Theory assumptions and proofs}
    \item[] Question: For each theoretical result, does the paper provide the full set of assumptions and a complete (and correct) proof?
    \item[] Answer: \answerYes{} 
    \item[] Justification: We analyze the theoretical communication volumes of different methods, and providing detailed derivation in Section \ref{subsec:communitation}.
    \item[] Guidelines:
    \begin{itemize}
        \item The answer NA means that the paper does not include theoretical results. 
        \item All the theorems, formulas, and proofs in the paper should be numbered and cross-referenced.
        \item All assumptions should be clearly stated or referenced in the statement of any theorems.
        \item The proofs can either appear in the main paper or the supplemental material, but if they appear in the supplemental material, the authors are encouraged to provide a short proof sketch to provide intuition. 
        \item Inversely, any informal proof provided in the core of the paper should be complemented by formal proofs provided in appendix or supplemental material.
        \item Theorems and Lemmas that the proof relies upon should be properly referenced. 
    \end{itemize}

    \item {\bf Experimental result reproducibility}
    \item[] Question: Does the paper fully disclose all the information needed to reproduce the main experimental results of the paper to the extent that it affects the main claims and/or conclusions of the paper (regardless of whether the code and data are provided or not)?
    \item[] Answer: \answerYes{} 
    \item[] Justification: We provide the experimental settings in Section \ref{subsec:settings}. We provide the pseudocode of our methods in Section \ref{sec:Implementation}.
    \item[] Guidelines:
    \begin{itemize}
        \item The answer NA means that the paper does not include experiments.
        \item If the paper includes experiments, a No answer to this question will not be perceived well by the reviewers: Making the paper reproducible is important, regardless of whether the code and data are provided or not.
        \item If the contribution is a dataset and/or model, the authors should describe the steps taken to make their results reproducible or verifiable. 
        \item Depending on the contribution, reproducibility can be accomplished in various ways. For example, if the contribution is a novel architecture, describing the architecture fully might suffice, or if the contribution is a specific model and empirical evaluation, it may be necessary to either make it possible for others to replicate the model with the same dataset, or provide access to the model. In general. releasing code and data is often one good way to accomplish this, but reproducibility can also be provided via detailed instructions for how to replicate the results, access to a hosted model (e.g., in the case of a large language model), releasing of a model checkpoint, or other means that are appropriate to the research performed.
        \item While NeurIPS does not require releasing code, the conference does require all submissions to provide some reasonable avenue for reproducibility, which may depend on the nature of the contribution. For example
        \begin{enumerate}
            \item If the contribution is primarily a new algorithm, the paper should make it clear how to reproduce that algorithm.
            \item If the contribution is primarily a new model architecture, the paper should describe the architecture clearly and fully.
            \item If the contribution is a new model (e.g., a large language model), then there should either be a way to access this model for reproducing the results or a way to reproduce the model (e.g., with an open-source dataset or instructions for how to construct the dataset).
            \item We recognize that reproducibility may be tricky in some cases, in which case authors are welcome to describe the particular way they provide for reproducibility. In the case of closed-source models, it may be that access to the model is limited in some way (e.g., to registered users), but it should be possible for other researchers to have some path to reproducing or verifying the results.
        \end{enumerate}
    \end{itemize}

\item {\bf Open access to data and code}
    \item[] Question: Does the paper provide open access to the data and code, with sufficient instructions to faithfully reproduce the main experimental results, as described in supplemental material?
    \item[] Answer: \answerYes{} 
    \item[] Justification: The link of Anonymous repository for ParaStep is provided in Section \ref{sec:Intro}.
    \item[] Guidelines:
    \begin{itemize}
        \item The answer NA means that paper does not include experiments requiring code.
        \item Please see the NeurIPS code and data submission guidelines (\url{https://nips.cc/public/guides/CodeSubmissionPolicy}) for more details.
        \item While we encourage the release of code and data, we understand that this might not be possible, so “No” is an acceptable answer. Papers cannot be rejected simply for not including code, unless this is central to the contribution (e.g., for a new open-source benchmark).
        \item The instructions should contain the exact command and environment needed to run to reproduce the results. See the NeurIPS code and data submission guidelines (\url{https://nips.cc/public/guides/CodeSubmissionPolicy}) for more details.
        \item The authors should provide instructions on data access and preparation, including how to access the raw data, preprocessed data, intermediate data, and generated data, etc.
        \item The authors should provide scripts to reproduce all experimental results for the new proposed method and baselines. If only a subset of experiments are reproducible, they should state which ones are omitted from the script and why.
        \item At submission time, to preserve anonymity, the authors should release anonymized versions (if applicable).
        \item Providing as much information as possible in supplemental material (appended to the paper) is recommended, but including URLs to data and code is permitted.
    \end{itemize}

\item {\bf Experimental setting/details}
    \item[] Question: Does the paper specify all the training and test details (e.g., data splits, hyperparameters, how they were chosen, type of optimizer, etc.) necessary to understand the results?
    \item[] Answer: \answerYes{} 
    \item[] Justification: This paper specify the test details in Section \ref{subsec:settings}. This paper does not include training.
    \item[] Guidelines:
    \begin{itemize}
        \item The answer NA means that the paper does not include experiments.
        \item The experimental setting should be presented in the core of the paper to a level of detail that is necessary to appreciate the results and make sense of them.
        \item The full details can be provided either with the code, in appendix, or as supplemental material.
    \end{itemize}

\item {\bf Experiment statistical significance}
    \item[] Question: Does the paper report error bars suitably and correctly defined or other appropriate information about the statistical significance of the experiments?
    \item[] Answer: \answerNo{} 
    \item[] Justification: The test datasets in our setting is big enough (946 samples for VBench) to perform evaluation, and we control the random seed for all of the models and methods used in our experiments.
    \item[] Guidelines:
    \begin{itemize}
        \item The answer NA means that the paper does not include experiments.
        \item The authors should answer "Yes" if the results are accompanied by error bars, confidence intervals, or statistical significance tests, at least for the experiments that support the main claims of the paper.
        \item The factors of variability that the error bars are capturing should be clearly stated (for example, train/test split, initialization, random drawing of some parameter, or overall run with given experimental conditions).
        \item The method for calculating the error bars should be explained (closed form formula, call to a library function, bootstrap, etc.)
        \item The assumptions made should be given (e.g., Normally distributed errors).
        \item It should be clear whether the error bar is the standard deviation or the standard error of the mean.
        \item It is OK to report 1-sigma error bars, but one should state it. The authors should preferably report a 2-sigma error bar than state that they have a 96\% CI, if the hypothesis of Normality of errors is not verified.
        \item For asymmetric distributions, the authors should be careful not to show in tables or figures symmetric error bars that would yield results that are out of range (e.g. negative error rates).
        \item If error bars are reported in tables or plots, The authors should explain in the text how they were calculated and reference the corresponding figures or tables in the text.
    \end{itemize}

\item {\bf Experiments compute resources}
    \item[] Question: For each experiment, does the paper provide sufficient information on the computer resources (type of compute workers, memory, time of execution) needed to reproduce the experiments?
    \item[] Answer: \answerYes{} 
    \item[] Justification: We provide the information on the computer resources in Section \ref{subsec:settings}.
    \item[] Guidelines:
    \begin{itemize}
        \item The answer NA means that the paper does not include experiments.
        \item The paper should indicate the type of compute workers CPU or GPU, internal cluster, or cloud provider, including relevant memory and storage.
        \item The paper should provide the amount of compute required for each of the individual experimental runs as well as estimate the total compute. 
        \item The paper should disclose whether the full research project required more compute than the experiments reported in the paper (e.g., preliminary or failed experiments that didn't make it into the paper). 
    \end{itemize}
    
\item {\bf Code of ethics}
    \item[] Question: Does the research conducted in the paper conform, in every respect, with the NeurIPS Code of Ethics \url{https://neurips.cc/public/EthicsGuidelines}?
    \item[] Answer: \answerYes{} 
    \item[] Justification: The research conducted in the paper conform with the NeurIPS Code of Ethics.
    \item[] Guidelines:
    \begin{itemize}
        \item The answer NA means that the authors have not reviewed the NeurIPS Code of Ethics.
        \item If the authors answer No, they should explain the special circumstances that require a deviation from the Code of Ethics.
        \item The authors should make sure to preserve anonymity (e.g., if there is a special consideration due to laws or regulations in their jurisdiction).
    \end{itemize}

\item {\bf Broader impacts}
    \item[] Question: Does the paper discuss both potential positive societal impacts and negative societal impacts of the work performed?
    \item[] Answer: \answerYes{}{} 
    \item[] Justification: We discuss that our methods can be a scalable and high-performance solution for acceleration diffusion model inference in practical environments in Appendix~\ref{sec:impacts}.
    \item[] Guidelines:
    \begin{itemize}
        \item The answer NA means that there is no societal impact of the work performed.
        \item If the authors answer NA or No, they should explain why their work has no societal impact or why the paper does not address societal impact.
        \item Examples of negative societal impacts include potential malicious or unintended uses (e.g., disinformation, generating fake profiles, surveillance), fairness considerations (e.g., deployment of technologies that could make decisions that unfairly impact specific groups), privacy considerations, and security considerations.
        \item The conference expects that many papers will be foundational research and not tied to particular applications, let alone deployments. However, if there is a direct path to any negative applications, the authors should point it out. For example, it is legitimate to point out that an improvement in the quality of generative models could be used to generate deepfakes for disinformation. On the other hand, it is not needed to point out that a generic algorithm for optimizing neural networks could enable people to train models that generate Deepfakes faster.
        \item The authors should consider possible harms that could arise when the technology is being used as intended and functioning correctly, harms that could arise when the technology is being used as intended but gives incorrect results, and harms following from (intentional or unintentional) misuse of the technology.
        \item If there are negative societal impacts, the authors could also discuss possible mitigation strategies (e.g., gated release of models, providing defenses in addition to attacks, mechanisms for monitoring misuse, mechanisms to monitor how a system learns from feedback over time, improving the efficiency and accessibility of ML).
    \end{itemize}
    
\item {\bf Safeguards}
    \item[] Question: Does the paper describe safeguards that have been put in place for responsible release of data or models that have a high risk for misuse (e.g., pretrained language models, image generators, or scraped datasets)?
    \item[] Answer: \answerNA{} 
    \item[] Justification: This paper provides a novel parallelism method for Diffusion models, does not pose such risks.
    \item[] Guidelines:
    \begin{itemize}
        \item The answer NA means that the paper poses no such risks.
        \item Released models that have a high risk for misuse or dual-use should be released with necessary safeguards to allow for controlled use of the model, for example by requiring that users adhere to usage guidelines or restrictions to access the model or implementing safety filters. 
        \item Datasets that have been scraped from the Internet could pose safety risks. The authors should describe how they avoided releasing unsafe images.
        \item We recognize that providing effective safeguards is challenging, and many papers do not require this, but we encourage authors to take this into account and make a best faith effort.
    \end{itemize}

\item {\bf Licenses for existing assets}
    \item[] Question: Are the creators or original owners of assets (e.g., code, data, models), used in the paper, properly credited and are the license and terms of use explicitly mentioned and properly respected?
    \item[] Answer: \answerYes{} 
    \item[] Justification: We cite the works used in this paper.
    \item[] Guidelines:
    \begin{itemize}
        \item The answer NA means that the paper does not use existing assets.
        \item The authors should cite the original paper that produced the code package or dataset.
        \item The authors should state which version of the asset is used and, if possible, include a URL.
        \item The name of the license (e.g., CC-BY 4.0) should be included for each asset.
        \item For scraped data from a particular source (e.g., website), the copyright and terms of service of that source should be provided.
        \item If assets are released, the license, copyright information, and terms of use in the package should be provided. For popular datasets, \url{paperswithcode.com/datasets} has curated licenses for some datasets. Their licensing guide can help determine the license of a dataset.
        \item For existing datasets that are re-packaged, both the original license and the license of the derived asset (if it has changed) should be provided.
        \item If this information is not available online, the authors are encouraged to reach out to the asset's creators.
    \end{itemize}

\item {\bf New assets}
    \item[] Question: Are new assets introduced in the paper well documented and is the documentation provided alongside the assets?
    \item[] Answer: \answerYes{} 
    \item[] Justification: The link of Anonymous repository for ParaStep is provided in Section \ref{sec:Intro}. This repository includes a README file, which provides instructions on installation, setup, and usage.
    \item[] Guidelines:
    \begin{itemize}
        \item The answer NA means that the paper does not release new assets.
        \item Researchers should communicate the details of the dataset/code/model as part of their submissions via structured templates. This includes details about training, license, limitations, etc. 
        \item The paper should discuss whether and how consent was obtained from people whose asset is used.
        \item At submission time, remember to anonymize your assets (if applicable). You can either create an anonymized URL or include an anonymized zip file.
    \end{itemize}

\item {\bf Crowdsourcing and research with human subjects}
    \item[] Question: For crowdsourcing experiments and research with human subjects, does the paper include the full text of instructions given to participants and screenshots, if applicable, as well as details about compensation (if any)? 
    \item[] Answer: \answerNA{} 
    \item[] Justification: This paper does not involve crowdsourcing nor research with human subjects.
    \item[] Guidelines:
    \begin{itemize}
        \item The answer NA means that the paper does not involve crowdsourcing nor research with human subjects.
        \item Including this information in the supplemental material is fine, but if the main contribution of the paper involves human subjects, then as much detail as possible should be included in the main paper. 
        \item According to the NeurIPS Code of Ethics, workers involved in data collection, curation, or other labor should be paid at least the minimum wage in the country of the data collector. 
    \end{itemize}

\item {\bf Institutional review board (IRB) approvals or equivalent for research with human subjects}
    \item[] Question: Does the paper describe potential risks incurred by study participants, whether such risks were disclosed to the subjects, and whether Institutional Review Board (IRB) approvals (or an equivalent approval/review based on the requirements of your country or institution) were obtained?
    \item[] Answer: \answerNA{} 
    \item[] Justification: This paper does not involve crowdsourcing nor research with human subjects.
    \item[] Guidelines:
    \begin{itemize}
        \item The answer NA means that the paper does not involve crowdsourcing nor research with human subjects.
        \item Depending on the country in which research is conducted, IRB approval (or equivalent) may be required for any human subjects research. If you obtained IRB approval, you should clearly state this in the paper. 
        \item We recognize that the procedures for this may vary significantly between institutions and locations, and we expect authors to adhere to the NeurIPS Code of Ethics and the guidelines for their institution. 
        \item For initial submissions, do not include any information that would break anonymity (if applicable), such as the institution conducting the review.
    \end{itemize}

\item {\bf Declaration of LLM usage}
    \item[] Question: Does the paper describe the usage of LLMs if it is an important, original, or non-standard component of the core methods in this research? Note that if the LLM is used only for writing, editing, or formatting purposes and does not impact the core methodology, scientific rigorousness, or originality of the research, declaration is not required.
    \item[] Answer: \answerNA{} 
    \item[] Justification: We only use LLM to assist writing.
    \item[] Guidelines:
    \begin{itemize}
        \item The answer NA means that the core method development in this research does not involve LLMs as any important, original, or non-standard components.
        \item Please refer to our LLM policy (\url{https://neurips.cc/Conferences/2025/LLM}) for what should or should not be described.
    \end{itemize}

\end{enumerate}

\end{document}